\documentclass[prodmode,acmcsur]{acmsmall} 

\usepackage[ruled]{algorithm2e}

\SetAlFnt{\small}
\SetAlCapFnt{\small}
\SetAlCapNameFnt{\small}
\SetAlCapHSkip{0pt}
\IncMargin{-\parindent}

\usepackage{amsmath}
\newcommand{\wa}{\mathbf{w}_a}
\newcommand{\wb}{\mathbf{w}_b}
\newcommand{\za}{\mathbf{z}_a}
\newcommand{\zb}{\mathbf{z}_b}
\newcommand{\bolda}{\boldsymbol\alpha}
\newcommand{\boldb}{\boldsymbol\beta}
\usepackage{comment}

\acmVolume{50}
\acmNumber{6}
\acmArticle{95}
\acmYear{2017}
\acmMonth{10}

\setcopyright{rightsretained}

\doi{10.1145/3136624}

\issn{0360-0300}

\begin{document}

\markboth{}{A Tutorial on Canonical Correlation Methods}

\title{A Tutorial on Canonical Correlation Methods}
\author{VIIVI UURTIO
\affil{Aalto University}
JO\~AO M. MONTEIRO
\affil{University College London}
JAZ KANDOLA
\affil{Imperial College London}
JOHN SHAWE-TAYLOR
\affil{University College London}
DELMIRO FERNANDEZ-REYES
\affil{University College London}
JUHO ROUSU
\affil{Aalto University}
}

\begin{abstract}
Canonical correlation analysis is a family of multivariate statistical methods for the analysis of paired sets of variables. Since its proposition, canonical correlation analysis has for instance been extended to extract relations between two sets of variables when the sample size is insufficient in relation to the data dimensionality, when the relations have been considered to be non-linear, and when the dimensionality is too large for human interpretation. This tutorial explains the theory of canonical correlation analysis including its regularised, kernel, and sparse variants. Additionally, the deep and Bayesian CCA extensions are briefly reviewed. Together with the numerical examples, this overview provides a coherent compendium on the applicability of the variants of canonical correlation analysis. By bringing together techniques for solving the optimisation problems, evaluating the statistical significance and generalisability of the canonical correlation model, and interpreting the relations, we hope that this article can serve as a hands-on tool for applying canonical correlation methods in data analysis.
\end{abstract}

%


%
%

\terms{Multivariate Statistical Analysis, Machine Learning, Statistical Learning Theory}


\acmformat{Viivi Uurtio, Jo\~ao M. Monteiro, Jaz Kandola, John Shawe-Taylor, Delmiro Fernandez-Reyes, and Juho Rousu, 2017. A Tutorial on Canonical Correlation Methods.}

\begin{bottomstuff}
Author's addresses: V. Uurtio (viivi.uurtio@aalto.fi) {and} J. Rousu (juho.rousu@aalto.fi), Helsinki Institute for Information Technology HIIT, Department of Computer Science, Aalto University, Konemiehentie 2,
02150 Espoo, Finland;
J. M. Monteiro (joao.monteiro@ucl.ac.uk), 
Department of Computer Science,
University College London,
and Max Planck Centre for Computational Psychiatry and Ageing Research,
University College London,
Gower Street, London WC1E 6BT, UK; 
J. Shawe-Taylor (j.shawe-taylor@ucl.ac.uk)  {and} D. Fernandez-Reyes (delmiro.fernandez-reyes@ucl.ac.uk),
Department of Computer Science,
University College London,
Gower Street, London WC1E 6BT, UK; 
J. Kandola (j.kandola@imperial.ac.uk), Division of Brain Sciences,
Imperial College London,
DuCane Road, London WC12 0NN.
\end{bottomstuff}

\maketitle

\section{Introduction}

When a process can be described by two sets of variables corresponding to two different aspects, or views, analysing the relations between these two views may improve the understanding of the underlying system. In this context, a relation is a mapping of the observations corresponding to a variable of one view to the observations corresponding to a variable of the other view. For example in the field of medicine, one view could comprise variables corresponding to the symptoms of the disease and the other to the risk factors that can have an effect on the disease incidence. Identifying the relations between the symptoms and the risk factors can improve the understanding of the disease exposure and give indications for prevention and treatment. Examples of these kind of two-view settings, where the analysis of the relations could provide new information about the functioning of the system, occur in several other fields of science. These relations can be determined by means of canonical correlation methods that have been developed specifically for this purpose.

Since the proposition of canonical correlation analysis (CCA) by H. Hotelling \cite{hotelling1935most,hotelling1936relations}, relations between variables have been explored in various fields of science. CCA was first applied to examine the relation of wheat characteristics to flour characteristics in an economics study by F. Waugh in 1942 \cite{waugh1942regressions}. Since then, studies in the fields of psychology \cite{hopkins1969statistical,dunham1975canonical}, geography \cite{monmonier1973improving}, medicine \cite{lindsey1985canonical}, physics \cite{wong1980study}, chemistry \cite{tu1989canonical}, biology \cite{sullivan1982distribution}, time-series modeling \cite{heij1991modified}, and signal processing \cite{schell1995programmable} constitute examples of the early application fields of CCA. 

In the beginning of the 21$^{st}$ century, the applicability of CCA has been demonstrated in modern fields of science such as neuroscience, machine learning and bioinformatics. Relations have been explored for developing brain-computer interfaces \cite{cao2015sequence,nakanishi2015comparison} and in the field imaging genetics \cite{fang2016joint}. CCA has also been applied for feature selection \cite{ogura2013variable}, feature extraction and fusion \cite{shen2013orthogonal}, and dimension reduction \cite{wang2013dimension}. Examples of application studies conducted in the fields of bioinformatics and computational biology include \cite{rousu2013biomarker,seoane2014canonical,baur2015canonical,sarkar2015dna,cichonska2016metacca}. The vast range of application domains emphasises the utility of CCA in extracting relations between variables. 

Originally, CCA was developed to extract linear relations in overdetermined settings, that is when the number of observations exceeds the number of variables in either view. To extend CCA to underdetermined settings that often occur in modern data analysis, methods of regularisation have been proposed. When the sample size is small, Bayesian CCA also provides an alternative to perform CCA. The applicability of CCA to underdetermined settings has been further improved through sparsity-inducing norms that facilitate the interpretation of the final result. Kernel methods and neural networks have been introduced for uncovering non-linear relations. At present, canonical correlation methods can be used to extract linear and non-linear relations in both over- and underdetermined settings.

In addition to the already described variants of CCA, alternative extensions have been proposed, such as the semi-paired and multi-view CCA. In general, CCA algorithms assume one-to-one correspondence between the observations in the views, in other words, the data is assumed to be paired. However, in real datasets some of the observations may be missing in either view, which means that the observations are semi-paired. Examples of semi-paired CCA algorithms comprise \cite{blaschko2008semi}, \cite{kimura2013semicca}, \cite{chen2012unified}, and \cite{zhang2014semi}. CCA has also been extended to more than two views by \cite{horst1961relations}, \cite{carroll1968generalization}, \cite{kettenring1971canonical}, and \cite{van1984linear}. In multi-view CCA the relations are sought among more than two views. Some of the modern extensions of multi-view CCA comprise its regularised \cite{tenenhaus2011regularized}, kernelised \cite{tenenhaus2015kernel}, and sparse \cite{tenenhaus2014variable} variants. Application studies of multi-view CCA and its modern variants can be found in neuroscience \cite{kang2013sparse}, \cite{chen2014removal}, feature fusion \cite{yuan2011novel} and dimensionality reduction \cite{yuan2014fractional}. However, both the semi-paired and multi-view CCA are beyond the scope of this tutorial. 

This tutorial begins with an introduction to the original formulation of CCA. The basic framework and statistical assumptions are presented. The techniques for solving the CCA optimisation problem are discussed. After solving the CCA problem, the approaches to interpret and evaluate the result are explained. The variants of CCA are illustrated using worked examples. Of the extended versions of CCA, the tutorial concentrates on the topics of regularised, kernel, and sparse CCA. Additionally, the deep and Bayesian CCA variants are briefly reviewed. This tutorial acquaints the reader with canonical correlation methods, discusses where they are applicable and what kind of information can be extracted.

\section{Canonical Correlation Analysis}

\subsection{The Basic Principles of CCA}\label{basic}

CCA is a two-view multivariate statistical method. In multivariate statistical analysis, the data comprises multiple variables measured on a set of observations or individuals. In the case of CCA, the variables of an observation can be partitioned into two sets that can be seen as the two views of the data. This can be illustrated using the following notations. Let the views $a$ and $b$ be denoted by the matrices $X_a$ and $X_b$, of sizes $n \times p$ and $n \times q$ respectively. The row vectors $\mathbf{x}_a^k \in \mathbb{R}^p$ and $\mathbf{x}_b^k \in \mathbb{R}^q$ for $k=1,2,\dots,n$ denote the sets of empirical multivariate observations in $X_a$ and $X_b$ respectively. The observations are assumed to be jointly sampled from a normal multivariate distribution. A reason for this is that the normal multivariate model approximates well the distribution of continuous measurements in several sampled distributions \cite{anderson1958introduction}. The column vectors $\mathbf{a}_i \in \mathbb{R}^n$ for $i=1,2,\dots,p$ and $\mathbf{b}_j \in \mathbb{R}^n$ for $j=1,2,\dots,q$ denote the variable vectors of the $n$ observations respectively. The inner product between two vectors is either denoted by $\langle \mathbf{a},\mathbf{b} \rangle$ or $\mathbf{a}^T \mathbf{b}$. Throughout this tutorial, we assume that the variables are standardised to zero mean and unit variance. In CCA, the aim is to extract the linear relations between the variables of $X_a$ and $X_b$.

CCA is based on linear transformations. We consider the following transformations
\begin{equation*}
X_a \wa = \za \quad \text{and } \quad X_b \wb = \zb
\end{equation*}
where $X_a \in \mathbb{R}^{n \times p}$, $\wa \in \mathbb{R}^p$, $\za \in \mathbb{R}^n$, $X_b \in \mathbb{R}^{n \times q}$, $\wb \in \mathbb{R}^q$, and $\zb \in \mathbb{R}^n$. The data matrices $X_a$ and $X_b$ represent linear transformations of the positions $\wa$ and $\wb$ onto the images $\za$ and $\zb$ in the space $\mathbb{R}^n$. The positions $\wa$ and $\wb$ are often referred to as canonical weight vectors and the images $\za$ and $\zb$ are also termed as canonical variates or scores. The constraints of CCA on the mappings are that the position vectors of the images $\za$ and $\zb$ are unit norm vectors and that the enclosing angle, $\theta \in [0, \frac{\pi}{2}]$ \cite{golub1995canonical,dauxois1997canonical}, between $\za$ and $\zb$ is minimised. The cosine of the angle, also referred to as the canonical correlation, between the images $\za$ and $\zb$ is given by the formula $\cos(\za,\zb) =  \langle \za,\zb \rangle / ||\za||||\zb||$ and due to the unit norm constraint $\cos(\za,\zb) = \langle \za,\zb \rangle.$ Hence the basic principle of CCA is to find two positions $\wa \in \mathbb{R}^p$ and $\wb \in \mathbb{R}^q$ that after the linear transformations $X_a \in \mathbb{R}^{n \times p}$ and $X_b \in \mathbb{R}^{n \times q}$ are mapped onto an $n$-dimensional unit ball and located in such a way that the cosine of the angle between the position vectors of their images $\za \in \mathbb{R}^n$ and $\zb \in \mathbb{R}^n$ is maximised.

The images $\za$ and $\zb$ of the positions $\wa$ and $\wb$ that result in the smallest angle, $\theta_1$, determine the first canonical correlation which equals $\cos \theta_1$ \cite{bjorck1973numerical}. The smallest angle is given by 
\begin{equation}
\label{alg_cca}
\begin{gathered} 
\cos \theta_1 = \max_{\za, \zb \in \mathbb{R}^n} \langle \za,\zb \rangle, \\
 ||\za||_2 = 1 \quad ||\zb||_2 = 1
\end{gathered}
\end{equation}
Let the maximum be obtained by $\za^1$ and $\zb^1$. The pair of images $\za^2$ and $\zb^2$, that has the second smallest enclosing angle $\theta_2$, is found in the orthogonal complements of $\za^1$ and $\zb^1$. The procedure is continued until no more pairs are found. Hence the $r$ angles $\theta_r \in [0, \frac{\pi}{2}]$ for $r=1,2,\cdots,q$ when $p>q$ that can be found are recursively defined by
\begin{gather*}
\cos \theta_r = \max_{\za, \zb \in \mathbb{R}^n} \langle \za^{r}, \zb^r \rangle, \\
 ||\za^r||_2 = 1 \quad ||\zb^r||_2 = 1 \\
\langle \za^{r}, \za^{j} \rangle =  0 \quad \langle \zb^{r}, \zb^{j} \rangle = 0, \\
\forall j \neq r: \quad j,r = 1, 2, \dots, \min(p,q).
\end{gather*}
The number of canonical correlations, $r$, corresponds to the dimensionality of CCA. Qualitatively, the dimensionality of CCA can be also seen as the number of patterns that can be extracted from the data.

When the dimensionality of CCA is large, it may not be relevant to solve all the positions $\wa$ and $\wb$ and images $\za$ and $\zb$. In general, the value of the canonical correlation and the statistical significance are considered to convey the importance of the pattern. The first estimation strategy for finding the number of statistically significant canonical correlation coefficients was proposed in \cite{bartlett1941statistical}. The techniques have been further developed in \cite{fujikoshi1979estimation,tu1991bootstrap,gunderson1997estimating,yamada2006permutation,lee2007canonical,sakurai2009asymptotic}.

In summary, the principle behind CCA is to find two positions in the two data spaces respectively that have images on a unit ball such that the angle between them is minimised and consequently the canonical correlation is maximised. The linear transformations of the positions are given by the data matrices. The number of relevant positions can be determined by analysing the values of the canonical correlations or by applying statistical significance tests.

\subsection{Finding the positions and the images in CCA}\label{solving}

The position vectors $\wa$ and $\wb$ having images $\za$ and $\zb$ in the new coordinate system of a unit ball that have a maximum cosine of the angle in between can be obtained using techniques of functional analysis. The eigenvalue-based methods comprise solving a standard eigenvalue problem, as originally proposed by Hotelling in \cite{hotelling1936relations}, or a generalised eigenvalue problem \cite{bach2002kernel,hardoon2004canonical}. Alternatively, the positions and the images can be found using the singular value decomposition (SVD), as introduced in \cite{healy1957rotation}. The techniques can be considered as standard ways of solving the CCA problem.

\paragraph{Solving CCA Through the Standard Eigenvalue Problem} In the technique of Hotelling, both the positions $\wa$ and $\wb$ and the images $\za$ and $\zb$ are obtained by solving a standard eigenvalue problem. The Lagrange multiplier technique \cite{hotelling1936relations,hooper1959simultaneous} is employed to obtain the characteristic equation. Let $X_a$ and $X_b$ denote the data matrices of sizes $n \times p$ and $n \times q$ respectively. The sample covariance matrix $C_{ab}$ between the variable column vectors in $X_a$ and $X_b$ is $C_{ab}=\frac{1}{n-1} X_a^T X_b$. The empirical variance matrices between the variables in $X_a$ and $X_b$ are given by $C_{aa}=\frac{1}{n-1}X_a^T X_a$ and $C_{bb}=\frac{1}{n-1} X_b^T X_b$ respectively. The joint covariance matrix is then
\begin{equation}\label{covmat}
\begin{pmatrix} C_{aa} & C_{ab} \\
C_{ba} & C_{bb}
\end{pmatrix}.
\end{equation}
The first and greatest canonical correlation that corresponds to the smallest angle is between the first pair of images $\mathbf{z}_a=X_a \wa$ and $\mathbf{z}_b=X_b \wb$. Since the correlation between $\mathbf{z}_a$ and $\mathbf{z}_b$ does not change with the scaling of $\mathbf{z}_a$ and $\mathbf{z}_b$, we can constrain $\wa$ and $\wb$ to be such that $\mathbf{z}_a$ and $\mathbf{z}_b$ have unit variance. This is given by 
\begin{align}
\mathbf{z}_a^T \mathbf{z}_a =& \wa^T X_a^T X_a \wa = \wa^T C_{aa} \wa = 1, \label{aa} \\ 
\mathbf{z}_b^T \mathbf{z}_b =& \wb^T X_b^T X_b \wb = \wb^T C_{bb} \wb = 1. \label{bb}
\end{align}
Due to the normality assumption and comparability, the variables of $X_a$ and $X_b$ should be centered such that they have zero means. In this case, the covariance between $\mathbf{z}_a$ and $\mathbf{z}_b$ is given by 
\begin{equation}\label{ab}
\mathbf{z}_a^T \mathbf{z}_b = \wa^T X_a^T X_b \wb = \wa^T C_{ab} \wb.
\end{equation}
Substituting (\ref{ab}), (\ref{aa}) and (\ref{bb}) into the algebraic problem in Equation (\ref{alg_cca}), we obtain:
\begin{gather*}
\cos \theta = \max_{\za, \zb \in \mathbb{R}^n} \langle \za,\zb \rangle = \max_{\wa \in \mathbb{R}^p, \wb \in \mathbb{R}^q} \wa^T C_{ab} \wb, \\
 ||\za||_2 = \sqrt{\wa^T C_{aa} \wa} = 1 \quad  ||\zb||_2 = \sqrt{\wb^T C_{bb} \wb} = 1.
\end{gather*}
In general, the constraints (\ref{aa}) and (\ref{bb}) are expressed in squared form,  $\wa^T C_{aa} \wa = 1$ and $\wb^T C_{bb} \wb = 1$. The problem can be solved using the Lagrange multiplier technique. Let 
\begin{equation}
L = \wa^T C_{ab} \wb - \frac{\rho_1}{2} (\wa^T C_{aa} \wa -1) - \frac{\rho_2}{2} (\wb^T C_{bb} \wb - 1)
\end{equation}
where $\rho_1$ and $\rho_2$ denote the Lagrange multipliers. Differentiating $L$ with respect to $\wa$ and $\wb$ gives
\begin{align}
\frac{\delta L}{\delta \wa} = C_{ab} \wb - \rho_1 C_{aa} \wa = \mathbf{0} \label{dif1}\\
\frac{\delta L}{\delta \wb} = C_{ba} \wa - \rho_2 C_{bb} \wb = \mathbf{0} \label{dif2}
\end{align}
Multiplying (\ref{dif1}) from the left by $\wa^T$ and (\ref{dif2}) from the left by $\wb^T$ gives
\begin{align*}
 \wa^T C_{ab} \wb - \rho_1 \wa^T C_{aa} \wa = 0 \\ 
\wb^T C_{ba} \wa - \rho_2 \wb^T C_{bb} \wb = 0. \\ 
\end{align*}
Since $\wa^T C_{aa} \wa = 1$ and $\wb^T C_{bb} \wb = 1$, we obtain that 
\begin{equation}\label{res}
\rho_1 = \rho_2 = \rho.
\end{equation}
Substituting (\ref{res}) into Equation (\ref{dif1}) we obtain
 \begin{equation}\label{wa}
 \wa = \frac{C_{aa}^{-1} C_{ab} \wb}{\rho}.
 \end{equation}
 Substituting (\ref{wa}) into (\ref{dif2}) we obtain
 \begin{equation*}
 \frac{1}{\rho} C_{ba} C_{aa}^{-1} C_{ab} \wb - \rho C_{bb} \wb = 0
 \end{equation*}
which is equivalent to the generalised eigenvalue problem of the form
\begin{equation*}
C_{ba} C_{aa}^{-1} C_{ab} \wb = \rho^2 C_{bb} \wb.
\end{equation*}
If $C_{bb}$ is invertible, the problem reduces to a standard eigenvalue problem of the form
\begin{equation*}
C_{bb}^{-1} C_{ba} C_{aa}^{-1} C_{ab} \wb = \rho^2 \wb.
\end{equation*}
The eigenvalues of the matrix $C_{bb}^{-1} C_{ba} C_{aa}^{-1} C_{ab}$ are found by solving the characteristic equation
\begin{equation*}
|C_{bb}^{-1} C_{ba} C_{aa}^{-1} C_{ab} - \rho^2 I | = 0.
\end{equation*} 
The square roots of the eigenvalues correspond to the canonical correlations. The technique of solving the standard eigenvalue problem is shown in Example \ref{cca_standard}. 

\begin{example}\label{cca_standard} We generate two data matrices $X_a$ and $X_b$ of sizes $n \times p$ and $n \times q$, where $n=60$, $p=4$ and $q=3$, respectively as follows. The variables of $X_a$ are generated from a random univariate normal distribution, $\mathbf{a}_1, \mathbf{a}_2, \mathbf{a}_3, \mathbf{a}_4 \sim N(0,1)$. We generate the following linear relations 
\begin{align*}
\mathbf{b}_1 &= \mathbf{a}_3 + \boldsymbol \xi_1 \\
\mathbf{b}_2 &= \mathbf{a}_1 + \boldsymbol \xi_2 \\
\mathbf{b}_3 &= -\mathbf{a}_4 + \boldsymbol \xi_3
\end{align*}
where $\boldsymbol \xi_1 \sim N(0,0.2), \boldsymbol \xi_2 \sim N(0,0.4),$ and $\boldsymbol \xi_3 \sim N(0,0.3)$ denote vectors of normal noise. The data is standardised such that every variable has zero mean and unit variance. The joint covariance matrix $C$ in (\ref{covmat}) of the generated data is given by
\begin{equation*}
C = 
\left(
\begin{array}{cccc|ccc}
1.00 & 0.34 & -0.11 & 0.21 & -0.10 & 0.92 & -0.21  \\
0.34 & 1.00 & -0.08 & 0.03 & -0.10 & 0.34 & 0.06  \\
-0.11 & -0.08 & 1.00 & -0.30 & 0.98 & -0.03 & 0.30  \\
0.21 & 0.03 & -0.30 & 1.00 & -0.25 & 0.12 & -0.94  \\
\hline
-0.10 & -0.10 & 0.98 & -0.25 & 1.00 & -0.03 & 0.25  \\
0.92 & 0.34 & -0.03 & 0.12 & -0.03 & 1.00 & -0.13  \\
-0.21 & 0.06 & 0.30 & -0.94 & 0.25 & -0.13  & 1.00 \\
\end{array}
\right)
=
\left(
\begin{array}{c|c}
C_{aa} & C_{ab}  \\
\hline
C_{ba} & C_{bb} \\
\end{array}
\right).
\end{equation*}
Now we compute the eigenvalues of the characteristic equation
\begin{equation*}
|C_{bb}^{-1} C_{ba} C_{aa}^{-1} C_{ab} - \rho^2 I| = 0.
\end{equation*}
The square roots of the eigenvalues of $ C_{bb}^{-1} C_{ba} C_{aa}^{-1} C_{ab}$ are $\rho_1 = 0.99$, $\rho_2 = 0.94$, and $\rho_3 = 0.92$. The eigenvectors $\wb$ satisfy the equation
\begin{equation*}
(C_{bb}^{-1} C_{ba} C_{aa}^{-1} C_{ab} - \rho^2 I) \wb = 0.
\end{equation*}
Hence we obtain
\begin{equation*}
\wb^1 = \begin{pmatrix}
-0.97 \\
-0.04 \\
-0.22 
\end{pmatrix}
\wb^2 = \begin{pmatrix}
-0.39 \\
-0.37 \\
0.85 
\end{pmatrix}
\wb^3 = \begin{pmatrix}
0.19 \\
-0.86 \\
-0.46 
\end{pmatrix}
\end{equation*}
and $\wa$ vectors satisfy
\begin{align*}
\wa^1 &= \frac{C_{aa}^{-1} C_{ab} \wb^1}{\rho_1} = 
\begin{pmatrix}
-0.04 \\
-0.00 \\
-0.99 \\
0.18
\end{pmatrix}
\wa^2 &= \frac{C_{aa}^{-1} C_{ab} \wb^2}{\rho_2} = 
\begin{pmatrix}
-0.41 \\
0.09 \\
-0.41 \\
-0.83
\end{pmatrix} 
\wa^3 &= \frac{C_{aa}^{-1} C_{ab} \wb^3}{\rho_3} = 
\begin{pmatrix}
-0.84 \\
-0.10 \\
0.14 \\
0.52 
\end{pmatrix}.
\end{align*}
The vectors $\wb^1,\wb^2$, and $\wb^3$ and $\wa^1,\wa^2$, and $\wa^3$ correspond to the pairs of positions $(\wa^1,\wb^1),(\wa^2,\wb^2)$ and $(\wa^3,\wb^3)$ that have the images $(\za^1,\zb^1),(\za^2,\zb^2)$ and $(\za^3,\zb^3)$. In linear CCA, the canonical correlations equal to the square roots of the eigenvalues, that is $\langle \za^1,\zb^1 \rangle = 0.99$, $\langle \za^2,\zb^2 \rangle = 0.94$, and $\langle \za^3,\zb^3 \rangle = 0.92$. $\qed$
\end{example}

\paragraph{Solving CCA Through the Generalised Eigenvalue Problem} The positions $\wa$ and $\wb$ and their images $\za$ and $\zb$ can also be solved through a generalised eigenvalue problem \cite{bach2002kernel,hardoon2004canonical}. The equations in (\ref{dif1}) and (\ref{dif2}) can be represented as simultaneous equations
\begin{align*}
C_{ab} \wb = & \rho C_{aa} \wa \\
C_{ba} \wa = & \rho C_{bb} \wb
\end{align*}
that are equivalent to 
\begin{equation}\label{geneig}
\begin{pmatrix}
\mathbf{0} & C_{ab} \\
C_{ba} & \mathbf{0}
\end{pmatrix}
\begin{pmatrix}
\wa \\
\wb
\end{pmatrix}=
\rho
\begin{pmatrix}
C_{aa} & \mathbf{0} \\
\mathbf{0} & C_{bb}
\end{pmatrix}
\begin{pmatrix}
\wa \\
\wb
\end{pmatrix}.
\end{equation}
The equation (\ref{geneig}) represents a generalised eigenvalue problem of the form $\beta A \mathbf{x} = \alpha B \mathbf{x}$ where the pair $(\beta,\alpha)=(1,\alpha)$ is an eigenvalue of the pair $(A,B)$ \cite{saad2011numerical,golub2012matrix}. The pair of matrices $A \in \mathbb{R}^{(p+q)\times(p+q)}$ and $B \in \mathbb{R}^{(p+q)\times(p+q)}$ is also referred to as matrix pencil. In particular, $A$ is symmetric and $B$ is symmetric positive-definite. The pair $(A,B)$ is then called the symmetric pair. As shown in \cite{watkins2004fundamentals}, a symmetric pair has real eigenvalues and $(p+q)$ linearly independent eigenvectors. To express the generalised eigenvalue problem in the form $A \mathbf{x} = \rho B \mathbf{x}$, the generalised eigenvalue is given by $\rho=\frac{\alpha}{\beta}$. Since the generalised eigenvalues come in pairs $\{\rho_1,-\rho_1,\rho_2,-\rho_2,\dots,\rho_p,-\rho_p,0\}$ where $p<q$, the positive generalised eigenvalues correspond to the canonical correlations. 

\begin{example}\label{cca_geneig} Using the data in Example \ref{cca_standard}, we apply the formulation of the generalised eigenvalue problem to obtain the positions $\wa$ and $\wb$. The resulting generalised eigenvalues are $$\{0.99,0.94,0.92,0.00,-0.92,-0.94,-0.99\}.$$ The generalised eigenvectors that correspond to the positive generalised eigenvalues in descending order are
\begin{equation*}
\wa^1  =
\begin{pmatrix}
-0.04  \\
-0.00 \\
-1.00 \\
0.18 
\end{pmatrix}
\wa^2  =
\begin{pmatrix}
0.48 \\
-0.11 \\
0.48 \\
0.98 
\end{pmatrix}
\wa^3  =
\begin{pmatrix}
-0.97  \\
-0.11 \\
0.16 \\
0.60 
\end{pmatrix}
\end{equation*} 
\begin{equation*}
\wb^1  =
\begin{pmatrix}
-0.98 \\
-0.04 \\
-0.23  
\end{pmatrix}
\wb^2  =
\begin{pmatrix}
0.46 \\
0.43 \\
-1.00  
\end{pmatrix}
\wb^3  =
\begin{pmatrix}
0.22 \\
-1.00 \\
-0.54  
\end{pmatrix}
\end{equation*} 
The vectors $\wa^1,\wa^2$, and $\wa^3$ and $\wb^1,\wb^2$, and $\wb^3$ correspond to the pairs of positions $(\wa^1,\wb^1),(\wa^2,\wb^2)$ and $(\wa^3,\wb^3).$  The canonical correlations are $\langle \za^1,\zb^1 \rangle = 0.99$, $\langle \za^2,\zb^2 \rangle = 0.94$, and $\langle \za^3,\zb^3 \rangle = 0.92$. 

The entries of the position pairs differ to some extent from the solutions to the standard eigenvalue problem in the Example \ref{cca_standard}. This is due to the numerical algorithms that are applied to solve the eigenvalues and eigenvectors. Additionally, the signs may also be opposite. This can be seen when comparing the second pairs of positions with the Example \ref{cca_standard}. This results from the symmetric nature of CCA. $\qed$
\end{example}

\paragraph{Solving CCA Using the SVD} The technique of applying the SVD to solve the CCA problem was first introduced by \cite{healy1957rotation} and described by \cite{ewerbring1989canonical} as follows. First, the variance matrices $C_{aa}$ and $C_{bb}$ are transformed into identity forms. Due to the symmetric positive definite property, the square root factors of the matrices can be found using a Cholesky or eigenvalue decomposition:
\begin{equation*}
C_{aa} = C_{aa}^{1/2} C_{aa}^{1/2} \quad \text{and} \quad C_{bb} = C_{bb}^{1/2} C_{bb}^{1/2}.
\end{equation*}
Applying the inverses of the square root factors symmetrically on the joint covariance matrix in (\ref{covmat}) we obtain
\begin{equation*}
\begin{pmatrix}  
C_{aa}^{-1/2} & \mathbf{0} \\
\mathbf{0} & C_{bb}^{-1/2} 
\end{pmatrix}
\begin{pmatrix}  
C_{aa} & C_{ab} \\
C_{ba} & C_{bb}
\end{pmatrix}
\begin{pmatrix}  
C_{aa}^{-1/2} & \mathbf{0} \\
\mathbf{0} & C_{bb}^{-1/2} 
\end{pmatrix}
=
\begin{pmatrix}  
I_{q} & C_{aa}^{-1/2} C_{ab} C_{bb}^{-1/2} \\
C_{bb}^{-1/2} C_{ba} C_{aa}^{-1/2} & I_{p} 
\end{pmatrix}.
\end{equation*}
The position vectors $\wa$ and $\wb$ can hence be obtained by solving the following SVD
\begin{equation}
C_{aa}^{-1/2} C_{ab} C_{bb}^{-1/2} = U^T S V
\end{equation}
where the columns of the matrices $U$ and $V$ correspond to the sets of orthonormal left and right singular vectors respectively. The singular values of matrix $S$ correspond to the canonical correlations. The positions $\wa$ and $\wb$ are obtained from 
\begin{equation*}
\wa  = C_{aa}^{-1/2} U \quad \wb  = C_{bb}^{-1/2} V
\end{equation*}
The method is shown in Example \ref{cca_svd}.

\begin{example}\label{cca_svd}
The method of solving CCA using the SVD is demonstrated using the data of Example \ref{cca_standard}. We compute the matrix
\begin{equation*}
C_{aa}^{-1/2} C_{ab} C_{bb}^{-1/2} =
\begin{pmatrix}
-0.02 & 0.90 & -0.06 \\
-0.07 & 0.20 & 0.11 \\
0.98 & 0.04 & 0.04 \\
0.01 & -0.02 & -0.93
\end{pmatrix}.
\end{equation*}
The SVD gives
\begin{gather*}
C_{aa}^{-1/2} C_{ab} C_{bb}^{-1/2} = \\
\underbrace{\begin{pmatrix}
-0.03 & -0.03 & 0.95 & -0.30  \\
-0.47 & 0.03 & -0.28 & 0.84 \\
-0.86 & -0.26 & 0.11 & 0.44 
\end{pmatrix}}_{U^T}
\underbrace{\begin{pmatrix}
0.99 & 0.00 & 0.00\\
0.00 & 0.94 & 0.00 \\
0.00 & 0.00 & 0.92 \\
0.00 & 0.00 & 0.00 
\end{pmatrix}}_{S}
\underbrace{\begin{pmatrix}
0.95 & -0.29 & 0.15 \\
0.01 & -0.44 & -0.90 \\
0.33 & 0.85 & -0.41
\end{pmatrix}}_{V}.
\end{gather*}
The singular values of the matrix $S$ correspond to the canonical correlations. The positions $\wa$ and $\wb$ are given by
\begin{equation*}
\wa^1 = C_{aa}^{-1/2} \mathbf{u}^1 =
\begin{pmatrix}
0.04  \\
0.00  \\
0.94  \\
-0.17 
\end{pmatrix}
\wa^2 = C_{aa}^{-1/2} \mathbf{u}^2 =
\begin{pmatrix}
-0.43  \\
0.10  \\
-0.43  \\
-0.87 
\end{pmatrix}
\wa^3 = C_{aa}^{-1/2} \mathbf{u}^3 =
\begin{pmatrix}
-0.91  \\
-0.10  \\
0.14  \\
0.56
\end{pmatrix}
\end{equation*}
\begin{equation*}
\wb^1  = C_{bb}^{-1/2} \mathbf{v}^1 =
\begin{pmatrix}
0.93  \\
0.04\\
0.21 
\end{pmatrix} 
\wb^2  = C_{bb}^{-1/2} \mathbf{v}^2 =
\begin{pmatrix}
-0.40  \\
-0.38 \\
0.89 
\end{pmatrix} 
\wb^3  = C_{bb}^{-1/2} \mathbf{v}^3 =
\begin{pmatrix}
0.21  \\
-0.93 \\
-0.50 
\end{pmatrix} 
\end{equation*}
where $\mathbf{u}^i$ and $\mathbf{v}^i$ for $i=1,2,3$ correspond to the left and right singular vectors. The vectors $\wa^1,\wa^2$, and $\wa^3$ and $\wb^1,\wb^2$, and $\wb^3$ correspond to the pairs of positions $(\wa^1,\wb^1),(\wa^2,\wb^2)$ and $(\wa^3,\wb^3).$ The canonical correlations are $\langle \za^1,\zb^1 \rangle = 0.99$, $\langle \za^2,\zb^2 \rangle = 0.94$, and $\langle \za^3,\zb^3 \rangle = 0.92$.  $\qed$
\end{example}


The main motivation for improving the eigenvalue-based technique was the computational complexity. The standard and generalised eigenvalue methods scale with the cube of the input matrix dimension, in other words, the time complexity is $\mathcal{O}(n^3)$, for a matrix of size $n \times n$. The input matrix $C_{aa}^{-1/2} C_{ab} C_{bb}^{-1/2}$ in the SVD-based technique is rectangular. This gives a time complexity of $\mathcal{O}(mn^2)$, for a matrix of size $m \times n$. Hence the SVD-based technique is computationally more tractable for very large datasets.

To recapitulate, the images $\za$ and $\zb$ of the positions $\wa$ and $\wb$ that successively maximise the canonical correlation can be obtained by solving a standard \cite{hotelling1936relations} or a generalised eigenvalue problem \cite{bach2002kernel,hardoon2004canonical} or by applying the SVD \cite{healy1957rotation,ewerbring1989canonical}. The CCA problem can also be solved using alternative techniques. The only requirements are that the successive images on the unit ball are orthogonal and that the angle is minimised.

\subsection{Evaluating the Canonical Correlation Model} 

The pair of position vectors that have images on the unit ball with a minimum enclosing angle correspond to the canonical correlation model obtained from the training data. The entries of these position vectors convey the relations between the variables obtained from the sampling distribution. In general, a statistical model is validated in terms of statistical significance and generalisability. To assess the statistical significance of the relations obtained from the training data, Bartlett's sequential test procedure \cite{bartlett1941statistical} can be applied. Although the technique was presented in 1941, it is still applied in timely CCA application studies such as \cite{marttinen2013genome,kabir2014canonical,song2016canonical}. The generalisability of the canonical correlation model determines whether the relations obtained from the training data can be considered to represent general patterns occurring in the sampling distribution. The methods of testing the statistical significance and generalisability of the extracted relations represent standard ways to evaluate the canonical correlation model.

\begin{figure}
\centerline{\includegraphics[trim={1cm 1cm 1.5cm 1cm},clip,width=0.78\textwidth]{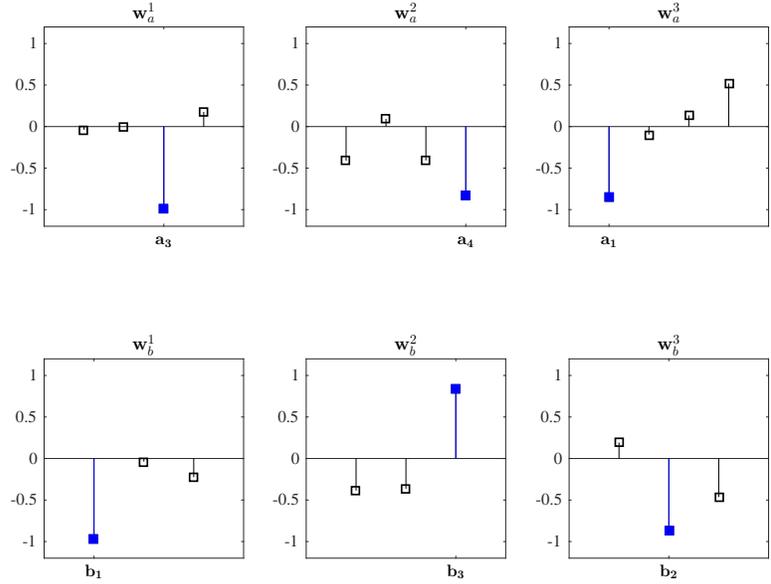}}
\caption{The entries of the pairs of positions $(\wa^1,\wb^1),(\wa^2,\wb^2)$ and $(\wa^3,\wb^3)$ are shown. The entry of maximum absolute value is coloured blue.}
\label{coefficients}
\end{figure}

The entries of the position vectors $\wa$ and $\wb$ can be used as a means to analyse the linear relations between the variables. The linear relation corresponding to the value of the canonical correlation is found between the entries that are of the greatest value. The values of the entries of the position vectors $\wa$ and $\wb$ are visualised in Figure \ref{coefficients}. The linear relation that corresponds to the canonical correlation of $\langle \za^1,\zb^1 \rangle=0.99$ is found between the variables $\mathbf{a}_3$ and $\mathbf{b}_1$. Since the signs of both entries are negative, the relation is positive. The second pair of positions $(\wa^2,\wb^2)$ conveys the negative relation between $\mathbf{a}_4$ and $\mathbf{b}_3$. The positive relation between $\mathbf{a}_1$ and $\mathbf{b}_2$ can be identified from the entries of the third pair of positions $(\wa^3,\wb^3)$.

In \cite{meredith1964canonical}, structure correlations were introduced as a means to analyse the relations between the variables. Structure correlations are the correlations of the original variables, $\mathbf{a}_i \in \mathbb{R}^n$ for $i=1,2,\dots,p$ and $\mathbf{b}_j \in \mathbb{R}^n$ for $j=1,2,\dots,q$, with the images, $\za \in \mathbb{R}^n$ or $\zb \in \mathbb{R}^n$. In general, the structure correlations convey how the images $\za$ and $\zb$ are aligned in the space $\mathbb{R}^n$ in relation to the variable axes. 

In \cite{ter1990interpreting}, the structure correlations were visualised on a biplot to facilitate the interpretation of the relations. To plot the variables on the biplot, the correlations of the original variables of both sets with two successive images, for example the images $(\za^1,\za^2)$, of one of the sets are computed. The plot is interpreted by the cosine of the angles between the variable vectors which is given by $\cos(\mathbf{a},\mathbf{b}) = \langle \mathbf{a},\mathbf{b} \rangle / ||\mathbf{a}||||\mathbf{b}||.$ Hence a positive linear relation is shown by an acute angle while an obtuse angle depicts a negative linear relation. A right angle corresponds to a zero correlation. 
\begin{figure}
\centerline{\includegraphics[trim={1cm 4cm 1.5cm 5cm},clip,width=0.78\textwidth]{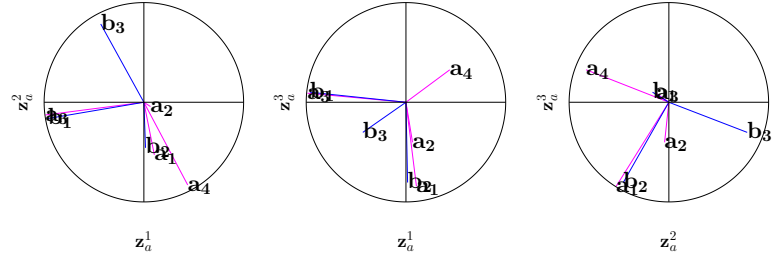}}
\caption{The biplots are generated using the results of Example \ref{cca_standard}. The biplot on the left shows the relations between the variables when viewed with respect to the images $\za^{1}$ and $\za^{2}$. The biplot in the middle shows the relations between the variables when viewed with respect to the images $\za^{1}$ and $\za^{3}$. The biplot on the right shows the relations between the variables when viewed with respect to the images $\za^{2}$ and $\za^{3}$. }
\label{biplot}
\end{figure}
Three biplots of the data and results of Example \ref{cca_standard} are shown in Figure \ref{biplot}. In each of the biplots, the same relations that were identified in Figure \ref{coefficients} can be found by analysing the angles between the variable vectors. The extraction of the relations can be enhanced by changing the pairs of images with which the correlations are computed.

The statistical significance tests of the canonical correlations evaluate whether the obtained pattern can be considered to occur non-randomly. The sequential test procedure of Bartlett \cite{bartlett1938further} determines the number of statistically significant canonical correlations in the data. The procedure to evaluate the statistical significance of the canonical correlations is described in \cite{fujikoshi1979estimation}. We test the hypothesis
\begin{equation}
H_0 : \min(p,q) = k \text{ against } H_1: \min(p,q) > k
\end{equation}
where $k = 0,1,\dots,p$ when $p<q$. If the hypothesis $H_0: \min(p,q) = j$ is rejected for $j = 0,1,\dots,k-1$ but accepted for $H_1: \min(p,q) > k-1$ the number of statistically significant canonical correlations can be estimated as $k$. For the test, the Bartlett-Lawley statistic, $L_k$ is applied
\begin{equation}
L_k = - \big(n-k-\frac{1}{2}(p+q+1) + \sum_{j=1}^{k} r_j^{-2} \big) \ln \big(\prod_{j=k+1}^{\min(p,q)} (1-r_j^2) \big).
\end{equation}
where $r_j$ denotes the $j^{th}$ canonical correlation. The asymptotic null distribution of $L_k$ is the chi-squared with $(p-k)(q-k)$ degrees of freedom. Hence we first test that no canonical relation exists between the two views. If we reject the hypothesis $H_0$ we continue to test that one canonical relation exists. If all the canonical patterns are statistically significant even the hypothesis $H_0 : \min(p,q) = k-1$ is rejected. 

\begin{example}\label{bartlett} We demonstrate the sequential test procedure of Bartlett using the simulated setting of Examples \ref{cca_standard}, \ref{cca_geneig} and \ref{cca_svd}. In the setting, $n=60$, $p=4$ and $p=3$. Hence $\min(p,q)=3$. First, we test that there are no canonical correlations
\begin{equation}
H_0 : \min(p,q) = 0 \text{ against } H_1: \min(p,q) > 0
\end{equation}
The Bartlett-Lawley statistic is $L_0 = 296.82$. Since $L_0 \sim \chi^2(12)$ the critical value at the significance level $\alpha = 0.01$ is $P(\chi^2 \geq 26.2) = 0.01$. Since $L_0 = 296.82 > 26.2$ the hypothesis $H_0$ is rejected. Next we test that there is one canonical correlation.
\begin{equation}
H_0 : \min(p,q) = 1 \text{ against } H_1: \min(p,q) > 1
\end{equation}
The Bartlett-Lawley statistic is $L_1 = 154.56$ and $L_1 \sim \chi^2(6)$. The critical value at the significance level $\alpha = 0.01$ is $P(\chi^2 \geq 16.8) = 0.01$. Since $L_1 = 154.56 > 16.8$ the hypothesis $H_0$ is rejected. We continue to test that there are two canonical correlations
\begin{equation}
H_0 : \min(p,q) = 2 \text{ against } H_1: \min(p,q) > 2
\end{equation}
The Bartlett-Lawley statistic is $L_2 = 70.95$ and $L_2 \sim \chi^2(2)$. The critical value at the significance level $\alpha = 0.01$ is $P(\chi^2 \geq 9.21) = 0.01$. Since $L_1 = 70.95 > 9.21$ the hypothesis $H_0$ is rejected. Hence the hypothesis $H_1: \min(p,q) > 2$ is accepted and all three canonical patterns are statistically significant.
$\qed$
\end{example}

To determine whether the extracted relations can be considered generalisable, or in other words general patterns in the sampling distribution, the linear transformations of the position vectors $\wa$ and $\wb$ need to be performed using test data. Unlike training data, test data originates from the sampling distribution but were not used in the model computation.  Let the matrices $X_{a}^{test} \in \mathbb{R}^{m \times p}$ and $X_{b}^{test} \in \mathbb{R}^{m \times q}$ denote the test data of $m$ observations. The linear transformations of the position vectors $\wa$ and $\wb$ are then
\begin{equation*}
X_a^{test} \wa = \za^{test} \quad \text{and } \quad X_b^{test} \wb = \zb^{test}
\end{equation*}
where the images $\za^{test}$ and $\zb^{test}$ are in the space $\mathbb{R}^m$. The cosine of the angle between the test images $\cos(\za^{test},\zb^{test})= \langle \za^{test}, \zb^{test} \rangle$ implies the generalisability. If the canonical correlations computed from test data also result in high correlation values we can deduce that the relations can generally be found from the particular sampling distribution.

\begin{example}\label{generalisability} We evaluate the generalisability of the canonical correlation model obtained in Example \ref{cca_standard}. The test data matrices $X_a^{test}$ and $X_b^{test}$ of sizes $m \times p$ and $m \times q$ where $m=40, p = 4,$ and $q=3$ are from the same distributions as described in Example \ref{cca_standard}. The $40$ observations were not included in the computation of the model. The test canonical correlations corresponding to the positions $(\wa^1,\wb^1),(\wa^2,\wb^2)$ and $(\wa^3,\wb^3)$  are $\langle \za^1,\zb^1 \rangle = 0.98$, $\langle \za^2,\zb^2 \rangle = 0.98$, $\langle \za^3,\zb^3 \rangle = 0.98.$ The high values indicate that the extracted relations can be considered generalisable.
$\qed$
\end{example}

The canonical correlation model can be evaluated by assessing the statistical significance and testing the generalisability of the relations. The statistical significance of the model can be determined by testing whether the extracted canonical correlations are not non-zero by chance. The generalisability of the relations can be assessed using new observations from the sampling distribution. These evaluation methods can generally be applied to test the validity of the extracted relations obtained using any variant of CCA.

\section{Extensions of Canonical Correlation Analysis}

\subsection{Regularisation Techniques in Underdetermined Systems}\label{reg}

CCA finds linear relations in the data when the number of observations exceeds the number of variables in either view. This possibly guarantees the non-singularity of the variance matrices $C_{aa}$ and $C_{bb}$ when solving the CCA problem. In the case of the standard eigenvalue problem, the matrices $C_{aa}$ and $C_{bb}$ should be non-singular so that they can be inverted. In the case of the SVD method, singular $C_{aa}$ and $C_{bb}$ may not have the square root factors. If the number of observations is less than the number of variables it is likely that some of the variables are collinear. Hence a sufficient sample size reduces the collinearity of the variables and guarantees the non-singularity of the variance matrices. The first proposition to solve the problem of insufficient sample size was presented in \cite{vinod1976canonical}. A more recent technique to regularise CCA has been proposed in \cite{cruz2014fast}. In the following, we present the original method of regularisation \cite{vinod1976canonical} due to its popularity in CCA applications \cite{gonzalez2009highlighting}, \cite{yamamoto2008canonical}, and \cite{soneson2010integrative}.

In the work of \cite{vinod1976canonical}, the singularity problem was proposed to be solved by regularisation. In general, the idea is to improve the invertibility of the variance matrices $C_{aa}$ and $C_{bb}$ by adding arbitrary constants $c_1 >  0$ and $c_2 >  0$ to the diagonal $C_{aa} + c_1 I$ and $C_{bb} + c_2 I.$ The constraints of CCA become
\begin{align*}
\wa^T \big(C_{aa} + c_1 I \big) \wa =& 1  \\ 
\wb^T \big(C_{bb} + c_2 I \big) \wb =& 1 
\end{align*}
and hence the magnitudes of the position vectors $\wa$ and $\wb$ are smaller when regularisation, $c_1 >  0$ and $c_2 >  0$, is applied. The regularised CCA optimisation problem is given by
\begin{gather*}
\cos \theta = \max_{\wa \in \mathbb{R}^p, \wb \in \mathbb{R}^q} \wa^T C_{ab} \wb, \\
\wa^T \big(C_{aa} + c_1 I \big) \wa = 1 \quad \wb^T \big(C_{bb} + c_2 I \big) \wb = 1.
\end{gather*}
The positions $\wa$ and $\wb$ can be found by solving the standard eigenvalue problem
\begin{equation*}
\big(C_{bb} + c_2 I \big)^{-1} C_{ba} \big(C_{aa} + c_1 I \big)^{-1} C_{ab} \wb = \rho^2 \wb.
\end{equation*}
or the generalised eigenvalue problem
\begin{equation*}\label{geneigreg}
\begin{pmatrix}
\mathbf{0} & C_{ab} \\
C_{ba} & \mathbf{0}
\end{pmatrix}
\begin{pmatrix}
\wa \\
\wb
\end{pmatrix}=
\rho
\begin{pmatrix}
C_{aa} + c_1 I & \mathbf{0} \\
\mathbf{0} & C_{bb} + c_2 I 
\end{pmatrix}
\begin{pmatrix}
\wa \\
\wb
\end{pmatrix}.
\end{equation*} 
As in the case of linear CCA, the canonical correlations correspond to the inner products between the consecutive image pairs $\langle \za^{i},\zb^{i} \rangle$ where $i=1,2,\dots,\min(p,q)$.

The regularisation proposed by \cite{vinod1976canonical} makes the CCA problem solvable but introduces new parameters $c_1 > 0$ and $c_2 > 0$ that have to be chosen. The first proposition of applying a leave-one-out cross-validation procedure to automatically select the regularisation parameters was presented in \cite{leurgans1993canonical}. Cross-validation is a well-established nonparametric model selection procedure to evaluate the validity of statistical predictions. One of its earliest applications have been presented in \cite{larson1931shrinkage}. A cross-validation procedure entails the partitioning of the observations into subsamples, selecting and estimating a statistic which is first measured on one subsample, and then validated on the other hold-out subsample. The method of cross-validation is discussed in detail for example in \cite{stone1974cross}, \cite{efron1979computers}, \cite{browne2000cross}, and more recently in \cite{arlot2010survey}. The cross-validation approach specifically developed for CCA has been further extended in \cite{waaijenborg2008quantifying,yamamoto2008canonical,gonzalez2009highlighting,soneson2010integrative}.  

In cross-validation, the size of the hold-out subsample varies depending on the size of the dataset. A leave-one-out cross-validation procedure is an option when the sample size is small and partitioning of the data into several folds, as is done in $k$-fold cross-validation, is not feasible. $5$-fold cross-validation saves computation time in relation to leave-one-out cross-validation if the sample size is large enough to partition the observations into five folds where each fold is used as a test set in turn. 

In general, as demonstrated for example in \cite{krstajic2014cross}, a $k$-fold cross-validation procedure should be repeated when an optimal set of parameters are searched for. Repetitions decrease the variance of the average values measured across the test folds. Algorithm \ref{alg:one} outlines an approach to determine the optimal regularisation parameters in CCA.
\begin{algorithm}[]
\SetAlgoNoLine
\KwIn{Data matrices $X_a$ and $X_b$, number of repetitions $R$, number of folds $F$}
\KwOut{Regularisation parameter values $c_1$ and $c_2$ maximising the correlation on test data}
Pre-defined ranges for values of $c_1$; $c_2$\;
Initialise $r=1$\;
\Repeat{$r=R$}{
        Randomly partition the observations into $F$ folds \;
        \For{all values of $c_1$ }
        {\For{all values of $c_2$}
        {\For{$i=1,2,\dots,F$}{
      Training set: $F-i$ folds, test set: $i$ fold\;
      Standardise the variables in the training and test sets\;
      For the training data, solve $|C_{bb}^{-1} C_{ba} \big(C_{aa} + c_1 I \big)^{-1} C_{ab} - \rho^2 I| = 0 $\;
      Find $\wb$ corresponding to the greatest eigenvalue satisfying $(C_{bb}^{-1} C_{ba} \big(C_{aa} + c_1 I \big)^{-1} C_{ab} - \rho^2 I) \wb = 0$ \;
      Find $\wa$ using $\wa^1 = \frac{\big(C_{aa} + c_1 I \big)^{-1} C_{ab} \wb^1}{\rho_1}$\;
      Transform the training positions $\wa$ and $\wb$ using the test observations $X_{a,test} \wa=\za$ and $X_{b,test} \wb=\zb$ \;
      Compute $\cos(\za,\zb) = \frac{\langle \za, \zb \rangle}{||\za||||\zb||}$\;
        }
        Store the mean of the $F$ values for $\cos(\za,\zb)$ obtained at $c_1$ and $c_2$\;
      }}
      $r=r+1$ \;}
      Compute the mean of the $R$ values for $\cos(\za,\zb)$ obtained at $c_1$ and $c_2$ \;
      Return the combination $c_1$ and $c_2$ that maximises $\cos(\za,\zb)$
\caption{Repeated k-fold cross-validation for regularised CCA}
\label{alg:one}
\end{algorithm}

\begin{figure}
\centerline{\includegraphics[trim={1cm 1cm 1.5cm 1cm},clip,width=0.78\textwidth]{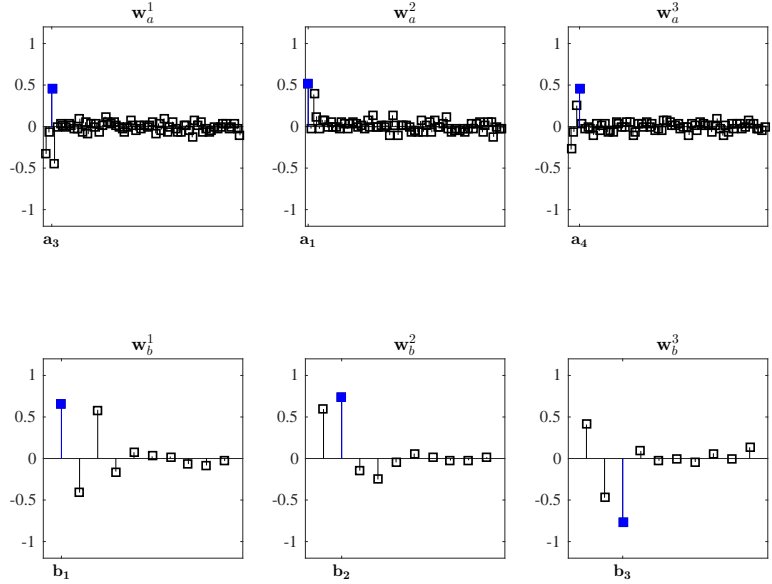}}
\caption{The entries of the pairs of positions $(\wa^1,\wb^1),(\wa^2,\wb^2)$ and $(\wa^3,\wb^3)$ are shown. The entry of maximum absolute value is coloured blue. The positive linear relation between $\mathbf{a}_3$ and $\mathbf{b}_1$, the positive linear relation between $\mathbf{a}_1$ and $\mathbf{b}_2$ and the negative linear relation between $\mathbf{a}_4$ and $\mathbf{b}_3$ are extracted by the pairs $(\wa^1,\wb^1)$, $(\wa^2,\wb^2)$, and $(\wa^3,\wb^3)$ respectively.}
\label{regcca}
\end{figure}

\begin{example}\label{reg_cca} To demonstrate the procedure of regularisation in underdetermined settings, we use the same simulated data as in the previous examples but we include additional normally distributed variables. The data matrices $X_a$ and $X_b$ of sizes $n \times p$ and $n \times q$, where $n=60$, $p=70$ and $q=10$, respectively as follows. The variables of $X_a$ are generated from a random univariate normal distribution, $\mathbf{a}_1, \mathbf{a}_2, \dots, \mathbf{a}_{70} \sim N(0,1)$. We generate the following linear relations 
\begin{align*}
\mathbf{b}_1 &= \mathbf{a}_3 + \boldsymbol \xi_1 \\
\mathbf{b}_2 &= \mathbf{a}_1 + \boldsymbol \xi_2 \\
\mathbf{b}_3 &= -\mathbf{a}_4 + \boldsymbol \xi_3
\end{align*}
where $\boldsymbol \xi_1 \sim N(0,0.01), \boldsymbol \xi_2 \sim N(0,0.03), \boldsymbol \xi_3 \sim N(0,0.02)$ denote vectors of normal noise. The remaining variables of $X_b$ are generated from random univariate normal distribution, $\mathbf{a}_4, \mathbf{a}_5, \dots, \mathbf{a}_{10} \sim N(0,1)$. The data is standardised such that every variable has zero mean and unit variance.

To construct the matrix $C_{bb}^{-1}C_{ba}C_{aa}^{-1}C_{ab}$, the variance matrices $C_{aa}$ and $C_{bb}$ need to be non-singular. Since $C_{aa}$ is obtained from a rectangular matrix, collinearity makes it close to singular. We therefore add a positive constant to the diagonal $C_{aa} + c_1 I$ to make it invertible. $C_{bb}$ is invertible since the data matrix $X_b$ has more rows than columns. The optimal value for the regularisation parameter $c_1$ can be determined for instance through repeated $k$-fold cross-validation. As shown in Figure \ref{test_cancor}, the optimal value $c_1 = 0.09$ was obtained through 50 times repeated 5-fold cross-validation using the procedure presented in the Algorithm \ref{alg:one}. 
\begin{figure}
\centerline{\includegraphics[trim={1.3cm 0cm 1.5cm 1cm},clip,width=0.38\textwidth]{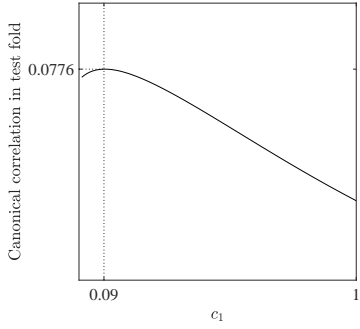}}
\caption{The maximum test canonical correlation, computed over 50 times repeated 5-fold cross-validation, is obtained at $c_1 = 0.09$.}
\label{test_cancor}
\end{figure}

The positions $\wa$ and $\wb$ and their respective images $\za$ and $\zb$ on a unit ball are found by solving the eigenvalues of the characteristic equation
\begin{equation}
|C_{bb}^{-1} C_{ba} \big(C_{aa} + c_1 I \big)^{-1} C_{ab} - \rho^2 I| = 0.
\end{equation}
The number of relations equals $min(p,q) = 10$. The square roots of the first three eigenvalues are $\rho_1 = 0.98$, $\rho_2 = 0.97$ and $\rho_3 = 0.96$. The respective three eigenvectors that correspond to the positions $\wb$ satisfy the equation
\begin{equation}
(C_{bb}^{-1} C_{ba} \big(C_{aa} + c_1 I \big)^{-1} C_{ab} - \rho^2 I) \wb = 0.
\end{equation}
The positions $\wa$ are found using the formula
\begin{equation}
\wa^i = \frac{\big(C_{aa} + c_1 I \big)^{-1} C_{ab} \wb^i}{\rho_i}
\end{equation}
where $i=1,2,3$ corresponds to the sorted eigenvalues and eigenvectors. By rounding correct to three decimal places, the first three canonical correlations are $\langle \za^1,\zb^1 \rangle = 0.999$, $\langle \za^2,\zb^2 \rangle = 0.998$, $\langle \za^3,\zb^3 \rangle = 0.996.$ The extracted linear relations are visualised in Figure \ref{regcca}. $\qed$
\end{example}

When either or both of the data views consists of more variables than observations, regularisation can be applied to make the variance matrices non-singular. This involves finding optimal non-negative scalar parameters that, when added to the diagonal entries, improve the invertibility of the variance matrices. After improving the invertibility of the variance matrices, the regularised CCA problem can be solved using the standard techniques.

\subsection{Bayesian Approaches for Robustness}\label{bayes}

Probabilistic approaches have been proposed to improve the robustness of CCA when the sample size is small and to be able to make more flexible distributional assumptions. A robust method generates a valid model regardless of outlying observations. In the following, a brief introduction to Bayesian CCA is provided. A detailed review on Bayesian CCA and its recent extensions can be found in \cite{klami2013bayesian}. 

An extension of CCA to probabilistic models was first proposed in \cite{bach2005probabilistic}. The probabilistic model contains the latent variables $\mathbf{y}^k \in \mathbb{R}^o$, where $o=\min(p,q)$, that generate the observations $\mathbf{x}_a^k \in \mathbb{R}^p$ and $\mathbf{x}_b^k \in \mathbb{R}^q$ for $k=1,2,\dots,n$. The latent variable model is defined by
\begin{equation*}
\begin{gathered}
\mathbf{y} \sim \mathcal{N}(0,I_d), \quad o \geq d \geq 1 \\
 \mathbf{x}_a | \mathbf{y} \sim \mathcal{N}(S_a\mathbf{y}+\boldsymbol{\mu}_a, \Psi_a), \quad S_a \in \mathbb{R}^{p \times d}, \Psi_a \succeq 0 \\
\mathbf{x}_b | \mathbf{y} \sim \mathcal{N}(S_b\mathbf{y}+\boldsymbol{\mu}_b, \Psi_b), \quad S_b \in \mathbb{R}^{q \times d}, \Psi_b \succeq 0 \\
\end{gathered}
\end{equation*}
where $\mathcal{N}(\boldsymbol{\mu},\Sigma)$ denotes the normal multivariate distribution with mean $\boldsymbol{\mu}$ and covariance $\Sigma$. The $S_a$ and $S_b$ correspond to the transformations of the latent variables $\mathbf{y}^k \in \mathbb{R}^o$. The $\Psi_a$ and $\Psi_b$ denote the noise covariance matrices. The maximum likelihood estimates of the parameters $S_a, S_b, \Psi_a, \Psi_b, \boldsymbol{\mu}_a$ and $\boldsymbol{\mu}_b$ are given by
\begin{equation*}
\begin{gathered}
\hat{S_a} = C_{aa} W_{ad} M_a \quad
\hat{S_b} = C_{bb} W_{bd} M_b \\ 
\hat{\Psi_a} = C_{aa}-\hat{S_a}\hat{S_a}^T \quad
\hat{\Psi_b} = C_{bb}-\hat{S_b}\hat{S_b}^T \\ 
\hat{\boldsymbol{\mu}}_a = \frac{1}{n} \sum_{k=1}^n \mathbf{x}_a^k \quad
\hat{\boldsymbol{\mu}}_b = \frac{1}{n} \sum_{k=1}^n \mathbf{x}_b^k \\
\end{gathered} 
\end{equation*}
where $M_a,M_b \in \mathbb{R}^{d \times d}$ are arbitrary matrices such that $M_a M_b^T=P_d$ and the spectral norms of $M_a$ and $M_b$ are smaller than one. $P_d$ is the diagonal matrix of the first $d$ canonical correlations. The $d$ columns of $W_{ad}$ and $W_{bd}$ correspond to the positions $\mathbf{w}_a^i$ and $\mathbf{w}_b^i$ for $i=1,2,\dots,d$ obtained using any of the standard techniques described in section \ref{basic}. 

The posterior expectations of $\mathbf{y}$ given $\mathbf{x}_a$ and $\mathbf{x}_b$ are $E(\mathbf{y}|\mathbf{x}_a)=M_a^T W_{ad}^T (\mathbf{x}_a-\hat{\mu_a})$ and $E(\mathbf{y}|\mathbf{x}_b)=M_b^T W_{bd}^T (\mathbf{x}_b-\hat{\mu_b})$. As stated in \cite{bach2005probabilistic}, regardless of what $M_a$ and $M_b$ are, $E(\mathbf{y}|\mathbf{x}_a)$ and $E(\mathbf{y}|\mathbf{x}_b)$ lie in the $d$-dimensional subspaces of $\mathbb{R}^p$ and  $\mathbb{R}^q$ which are identical to those obtained by linear CCA. The generative model of \cite{bach2005probabilistic} was further developed in \cite{archambeau2006robust} by replacing the normal noise with the multivariate Student's t distribution. This improves the robustness against outlying observations that are then better modeled by the noise term \cite{klami2013bayesian}.

A Bayesian extension of CCA was proposed by \cite{klami2007local} and \cite{wang2007variational}. To perform Bayesian analysis, the probabilistic model has to be supplemented with prior distributions of the model parameters. In \cite{klami2007local} and \cite{wang2007variational}, the prior distribution of the covariance matrices $\Psi_a$ and $\Psi_b$ was chosen to be the inverse-Wishart distribution. The automatic relevance determination \cite{neal2012bayesian} prior was selected for the linear transformations $S_a$ and $S_b$. The inference on the posterior distribution was made by applying a variational mean-field algorithm \cite{wang2007variational} and Gibbs sampling \cite{klami2007local}.

As in the case of the linear CCA, the variance matrices obtained from high-dimensional data make the inference of the probabilistic and Bayesian CCA models difficult \cite{klami2013bayesian}. This is because the variance matrices need to be inverted in the inference algorithms. To perform Bayesian CCA on high-dimensional data, dimensionality reduction techniques should be applied as a preprocessing step, as has been done for example in \cite{huopaniemi2010multivariate}. 

An advantage of Bayesian CCA, in relation to linear CCA, is the application of the prior distributions that enable to take the possible underlying structure in the data into account. Examples of studies where sparse models were obtained by means of the prior distribution include \cite{archambeau2009sparse} and \cite{rai2009multi}. In addition to modeling the structure of the data, in \cite{klami2012bayesian} the Bayesian CCA was extended such that any exponential family distribution could model the noise, not only the normal. 

In summary, probabilistic and Bayesian CCA provide alternative ways to interpret the CCA by means of latent variables. Bayesian CCA may be more feasible in settings where knowledge regarding the data can be incorporated through the prior distributions. Additionally, noise can be modelled by other exponential family distribution functions than the normal distribution.

\subsection{Uncovering Linear and Non-Linear Relations}

CCA \cite{hotelling1936relations} finds linear relations between variables belonging to two views that both are overdetermined. The first proposition to extend CCA to uncover non-linear relations using an optimal scaling method was presented in \cite{burg1983non}. At the turn of the 21$^{st}$ century, artificial neural networks were incorporated in the CCA framework for finding non-linear relations \cite{lai1999neural,fyfe2000canonical,hsieh2000nonlinear}. Deep CCA \cite{andrew2013deep} is an example of a recent non-linear CCA variant employing artificial neural networks. Shortly after the introduction of the neural networks, propositions of applying kernel methods in CCA were presented in \cite{lai2000kernel,akaho2001kernel,van2001kernel,melzer2001nonlinear,bach2002kernel}. Since then, the kernelised version of CCA has received considerable attention in terms of theoretical foundations \cite{hardoon2004canonical,fukumizu2007statistical,alam2008sensitivity,blaschko2008semi,hardoon2009convergence,cai2013distance} and applications \cite{melzer2003appearance,wang2005nonlinear,hardoon2007unsupervised,larson2014kernel}. In the following, we present how kernel CCA can be applied to uncover nonlinear relations between the variables. We then provide a brief overview on deep CCA.

To extract linear relations, CCA is performed in the data spaces of $X_a \in \mathbb{R}^{n \times p}$ and $X_b \in \mathbb{R}^{n \times q}$ where the $n$ rows correspond to the observations and the $p$ and $q$ columns correspond to the variables. The relations between the variables are found analysing the positions $\wa \in \mathbb{R}^p$ and $\wb \in \mathbb{R}^q$ that have such images $\za = X_a \wa$ and $\zb = X_b \wb$ on a unit ball in $\mathbb{R}^n$ that have a minimum enclosing angle. The extracted relations are linear since the positions $\wa$ and $\wb$ and their images $\za$ and $\zb$ were obtained in the Euclidean space.

To extract non-linear relations, the positions $\wa$ and $\wb$ should be found in a space where the distances, or measures of similarity, between objects are non-linear. This can be achieved using kernel methods, that is by transforming the original observations $\mathbf{x}_a^i \in \mathbb{R}^p $ and $\mathbf{x}_b^i \in \mathbb{R}^q$, where $i=1,2,\dots,n$, to Hilbert spaces $\mathcal{H}_a$ and $\mathcal{H}_b$ through feature maps $\phi_a: \mathbb{R}^p \mapsto \mathcal{H}_a$ and $\phi_b: \mathbb{R}^q \mapsto \mathcal{H}_b$. The similarity of the objects is captured by a symmetric positive semi-definite kernel, corresponding to the inner products in the Hilbert spaces  
$K_a(\mathbf{x}^i_a,\mathbf{x}^j_a) = \langle \boldsymbol{\phi}_a(\mathbf{x}^i_a),\boldsymbol{\phi}_a(\mathbf{x}^j_a) \rangle_{\mathcal{H}_a}$ and $K_b(\mathbf{x}^i_b,\mathbf{x}^j_b) = \langle \boldsymbol{\phi}_b(\mathbf{x}^i_b),\boldsymbol{\phi}_b(\mathbf{x}^j_b)\rangle_{\mathcal{H}_b}.$ The feature maps are typically non-linear and result in high-dimensional intrinsic spaces $\boldsymbol{\phi}_a(\mathbf{x}_a^i) \in \mathcal{H}_a$ and $\boldsymbol{\phi}_b(\mathbf{x}_b^i) \in \mathcal{H}_b$ for $i=1,2,\dots,n$. Through kernels, CCA can be used to extract non-linear correlations, relying on the fact that the CCA solution can always be found within the span of the data \cite{bach2002kernel,scholkopf1998nonlinear}. 

The basic principles behind kernel CCA are similar to CCA. First, the observations are transformed to Hilbert spaces $\mathcal{H}_a$ and $\mathcal{H}_b$ using symmetric positive semi-definite kernels
\begin{equation*}
K_a(\mathbf{x}^i_a,\mathbf{x}^j_a) = \langle \boldsymbol{\phi}_a(\mathbf{x}^i_a),\boldsymbol{\phi}_a(\mathbf{x}^j_a) \rangle_{\mathcal{H}_a} \text{ and } K_b(\mathbf{x}^i_b,\mathbf{x}^j_b) = \langle \boldsymbol{\phi}_b(\mathbf{x}^i_b),\boldsymbol{\phi}_b(\mathbf{x}^j_b)\rangle_{\mathcal{H}_b}
\end{equation*}
where $i,j=1,2,\dots,n$. As derived in \cite{bach2002kernel}, the original data matrices $X_a \in \mathbb{R}^{n \times p}$ and $X_b \in \mathbb{R}^{n \times q}$ can be substituted by the Gram matrices $K_a \in \mathbb{R}^{n \times n}$ and $K_b \in \mathbb{R}^{n \times n}$. Let $\bolda$ and $\boldb$ denote the positions in the kernel space $\mathbb{R}^n$ that have the images $\za=K_a \bolda$ and $\zb=K_b \boldb$ on the unit ball in $\mathbb{R}^n$ with a minimum enclosing angle in between. The kernel CCA problem is hence
\begin{gather}
\cos(\za,\zb) = \max_{\za, \zb \in \mathbb{R}^n} \langle \za,\zb \rangle = \bolda^T K_a^T K_b \boldb, \\
 ||\za||_2=\sqrt{\bolda^T K_a^2 \bolda} = 1 \quad ||\zb||_2=\sqrt{\boldb^T K_b^2 \boldb} = 1
\end{gather}
As in CCA, the optimisation problem can be solved using the Lagrange multiplier technique.
\begin{equation}
L = \bolda^T K_a^T K_b \boldb - \frac{\rho_1}{2} (\bolda^T K_a^2 \bolda -1) - \frac{\rho_2}{2} (\boldb^T K_b^2 \boldb - 1)
\end{equation}
where $\rho_1$ and $\rho_2$ denote the Lagrange multipliers. Differentiating $L$ with respect to $\bolda$ and $\boldb$ gives
\begin{align}
\frac{\delta L}{\delta \bolda} = K_a K_b \boldb - \rho_1 K_a^2 \bolda = \mathbf{0} \label{kdif1}\\
\frac{\delta L}{\delta \boldb} = K_b K_a \bolda - \rho_2 K_b^2 \boldb = \mathbf{0} \label{kdif2}
\end{align}
Multiplying (\ref{dif1}) from the left by $\bolda^T$ and (\ref{dif2}) from the left by $\boldb^T$ gives
\begin{align}
\bolda^T K_a K_b \boldb - \rho_1 \bolda^T K_a^2 \bolda = 0 \label{ks1}\\
\boldb^T K_b K_a \bolda - \rho_2 \boldb^T K_b^2 \boldb = 0. \label{ks2}
\end{align}
Since $\bolda^T K_a^2 \bolda = 1$ and $\boldb^T K_b^2 \boldb = 1$, we obtain that 
\begin{equation}\label{kres}
\rho_1 = \rho_2 = \rho.
\end{equation}
Substituting (\ref{kres}) into Equation (\ref{kdif1}) we obtain
 \begin{equation}\label{kwa}
 \bolda = \frac{K_a^{-1} K_a^{-1} K_a K_b \boldb}{\rho}=\frac{K_a^{-1} K_b \boldb}{\rho}.
 \end{equation}
 Substituting (\ref{kwa}) into (\ref{kdif2}) we obtain
 \begin{equation}
 \frac{1}{\rho} K_b K_a K_a^{-1} K_b \boldb - \rho K_b^2 \boldb = 0
 \end{equation}
which is equivalent to the generalised eigenvalue problem of the form
\begin{equation}
K_b^2 \boldb = \rho^2 K_b^2 \boldb.
\end{equation}
If $K_b^2$ is invertible, the problem reduces to a standard eigenvalue problem of the form
\begin{equation}
I \boldb = \rho^2 \boldb.
\end{equation}
Clearly, in the kernel space, if the Gram matrices are invertible the resulting canonical correlations are all equal to one. Regularisation is therefore needed to solve the kernel CCA problem.

Kernel CCA can be regularised in a similar manner as presented in Section \ref{reg} \cite{bach2002kernel,hardoon2004canonical}. We constrain the norms of the position vectors $\bolda$ and $\boldb$ by adding constants $c_1$ and $c_2$ to the diagonals of the Gram matrices $K_a$ and $K_b$ 
\begin{align}
\bolda^T \big(K_a + c_1 I \big)^2 \bolda =& 1 \label{kvara} \\ 
\boldb^T \big(K_b + c_2 I \big)^2 \boldb =& 1. \label{kvarb}
\end{align}
The solution can then be found by solving the standard eigenvalue problem 
\begin{equation*}
\big(K_b + c_1 I \big)^{-2} K_b K_a \big(K_a + c_2 I \big)^{-2} K_a K_b \bolda = \rho^2 \bolda.
\end{equation*}
As in the case of CCA, kernel CCA can also be solved through the generalised eigenvalue problem \cite{bach2002kernel}. Since the data matrices $X_a$ and $X_b$ can be substituted by the corresponding Gram matrices $K_a$ and $K_b$, the formulation becomes
\begin{equation}\label{kgeneig}
\begin{pmatrix}
\mathbf{0} & K_a K_b \\
K_b K_a & \mathbf{0}
\end{pmatrix}
\begin{pmatrix}
\bolda \\
\boldb
\end{pmatrix}=
\rho
\begin{pmatrix}
\big(K_a + c_1 I \big)^2 & \mathbf{0} \\
\mathbf{0} & \big(K_b + c_2 I \big)^2
\end{pmatrix}
\begin{pmatrix}
\bolda \\
\boldb
\end{pmatrix}
\end{equation}
where the constants $c_1$ and $c_2$ denote the regularisation parameters. In Example \ref{kcca_ex}, kernel CCA, solved through the generalised eigenvalue problem, is performed on simulated data.

\begin{example}\label{kcca_ex} We generate a simulated dataset as follows. The data matrices $X_a$ and $X_b$ of sizes $n \times p$ and $n \times q$, where $n=150$, $p=7$ and $q=8$, respectively as follows. The seven variables of $X_a$ are generated from a random univariate normal distribution, $\mathbf{a}_1, \mathbf{a}_2, \dots, \mathbf{a}_7 \sim N(0,1)$. We generate the following relations 
\begin{align*}
\mathbf{b}_1 &= \exp(\mathbf{a}_3) + \boldsymbol \xi_1 \\
\mathbf{b}_2 &= \mathbf{a}_1^3 + \boldsymbol \xi_2 \\
\mathbf{b}_3 &= -\mathbf{a}_4 + \boldsymbol \xi_3
\end{align*}
where $\boldsymbol \xi_1 \sim N(0,0.4)$, $\boldsymbol \xi_2 \sim N(0,0.2)$ and $\boldsymbol \xi_3 \sim N(0,0.3)$ denote vectors of normal noise. The five other variables of $X_b$ are generated from a random univariate normal distribution, $\mathbf{b}_4, \mathbf{b}_5, \dots, \mathbf{b}_8 \sim N(0,1)$. The data is standardised such that every variable has zero mean and unit variance.

In kernel CCA, the choice of the kernel function affects what kind of relations can be extracted. In general, a Gaussian kernel $K(\mathbf{x},\mathbf{y})=exp(-\frac{||\mathbf{x}-\mathbf{y}||^2}{2\sigma^2})$ is used when the data is assumed to contain non-linear relations. The width parameter $\sigma$ determines the non-linearity in the distances between the data points computed in the form of inner products. Increasing the value of $\sigma$ makes the space closer to Euclidean while decreasing makes the distances more non-linear. The optimal value for $\sigma$ is best determined using a re-sampling method such as a cross-validation scheme, for example procedure similar to the one presented in Algorithm \ref{alg:one}. In this example, we applied the "median trick", presented in \cite{song2010hilbert}, according to which the $\sigma$ corresponds to the median of Euclidean distances computed between all pairs of observations. The median distances for the data in this example were $\sigma_a = 3.53 $ and $\sigma_b = 3.62$ for the views $X_a$ and $X_b$ respectively. The kernels were centred by $\tilde{K} = K - \frac{1}{n} \mathbf{j} \mathbf{j}^T K - \frac{1}{n} K \mathbf{j} \mathbf{j}^T  + \frac{1}{n^2} (\mathbf{j}^T K\mathbf{j}) \mathbf{j} \mathbf{j}^T $ where $\mathbf{j}$ contains only entries of value one \cite{shawe2004kernel}.

In addition to the kernel parameters, also the regularisation parameters $c_1$ and $c_2$ need to be optimised to extract the correct relations. As in the case of regularised CCA, a repeated cross-validation procedure can be applied to identify the optimal pair of parameters. For the data in this example, the optimal regularisation parameters were $c_1=1.50$ and $c_2=0.60$ when a 20 times repeated 5-fold cross-validation was applied. The first three canonical correlations at the optimal parameter values were $\langle \za^1,\zb^1 \rangle = 0.95$, $\langle \za^2,\zb^2 \rangle=0.89$, and $\langle \za^3,\zb^3 \rangle=0.87.$

The interpretation of the relations cannot be performed from the positions $\bolda$ and $\boldb$ since they are obtained in the kernel spaces. In the case of simulated data, we know what kind of relations are contained in the data. We can compute the linear correlation coefficient between the simulated relations and the transformed pairs of positions $\za$ and $\zb$ \cite{chang2013canonical}. The correlation coefficients are shown in Table \ref{tab:one}. The exponential relation was extracted in the second pair $(\za^2,\zb^2)$, the 3$^{rd}$ order polynomial relation was extracted in the third pair $(\za^3,\zb^3)$ and the linear relation in the first pair $(\za^1,\zb^1)$.
\begin{table}%
\tbl{Extracted relations by kernel CCA \label{tab:one}}{%
\begin{tabular}{|c|c|c|c|}
\hline
  & $\za^1$ & $\za^2$ & $\za^3$ \\\hline
$exp(\mathbf{a}_3)$  & 0.00 & \bf{0.81} & 0.09  \\\hline
$\mathbf{a}_1^3$  & 0.05 & 0.14 & \bf{0.74} \\\hline
$-\mathbf{a}_4$  & \bf{0.99} & 0.07 & 0.04  \\\hline
  & $\zb^1$ & $\zb^2$ & $\zb^3$ \\\hline
$\mathbf{b}_1$  & 0.02 & \bf{0.93} & 0.12 \\\hline
$\mathbf{b}_2$   & 0.08 & 0.15 & \bf{0.87} \\\hline
$\mathbf{b}_3$  & \bf{0.98} & 0.01 & 0.03 \\\hline
\end{tabular}}
\begin{tabnote}%
\end{tabnote}%
\end{table}%
 $\qed$
\end{example}

In \cite{hardoon2004canonical}, an alternative formulation of the standard eigenvalue problem was presented when the data contains a large number of observations. If the sample size is large, the dimensionality of the Gram matrices $K_a$ and $K_b$ can cause computational problems. Partial Gram-Schmidt orthogonalization (PGSO) \cite{cristianini2002latent} was proposed as a matrix decomposition method. PGSO results in
\begin{align*}
K_a \simeq& R_a R_a^T  \\
K_b \simeq& R_bR_b^T.
\end{align*}
Substituting these into the Equations (\ref{kdif1}) and (\ref{kdif2}) and multiplying by $R_a^T$ and $R_b^T$ respectively we obtain 
\begin{align}
R_a^T R_a R_a^TR_b R_b^T\boldb - \rho R_a^T R_a^T R_a R_a^T R_a \bolda = \mathbf{0} \label{4.1}\\
R_b^T R_b R_b^T R_a R_a^T \bolda - \mu R_b^T R_b^T R_b R_b^T R_b \boldb = \mathbf{0} \label{4.2}.
\end{align}
Let $D_{aa}=R_a^TR_a$, $D_{ab}=R_a^TR_b$, $D_{ba}=R_b^TR_a$, and $D_{bb}=R_b^TR_b$ denote the blocks of the new sample covariance matrix. Let $\tilde{\bolda}=R_a^T\bolda$ and $\tilde{\boldb}=R_b^T\boldb$ denote the positions $\bolda$ and $\boldb$ in the reduced space. Using these substitutions in (\ref{4.1}) and (\ref{4.2}) we obtain
\begin{align}
D_{aa}D_{ab}\tilde{\boldb} - \rho D_{aa}^2 \tilde{\bolda} = \mathbf{0} \label{4.3}\\
D_{bb}D_{ba}\tilde{\bolda} - \rho D_{bb}^2 \tilde{\boldb} = \mathbf{0} \label{4.4}.
\end{align}
If $D_{aa}$ and $D_{bb}$ are invertible we can multiply (\ref{4.3}) by $D_{aa}^{-1}$ and (\ref{4.4}) by $D_{bb}^{-1}$  which gives
\begin{align}
D_{ab}\tilde{\boldb} - \rho D_{aa} \tilde{\bolda} = \mathbf{0} \label{4.5}\\
D_{ba}\tilde{\bolda} - \rho D_{bb} \tilde{\boldb} = \mathbf{0} \label{4.6}.
\end{align}
and hence 
\begin{equation}\label{hardoon_beta}
\tilde{\boldb} = \frac{D_{bb}^{-1} D_{ba} \tilde{\bolda}}{\rho}
\end{equation}
which, after a substitution into (\ref{4.3}), results in a generalised eigenvalue problem 
\begin{equation}\label{kgene}
D_{ab}D_{bb}^{-1} D_{ba} \tilde{\bolda} = \rho^2 D_{aa} \tilde{\bolda}.
\end{equation}
To formulate the problem as a standard eigenvalue problem, let $D_{aa}=SS^T$ denote the complete Cholesky decomposition where $S$ is a lower triangular matrix and let $\hat{\bolda}=S^T\tilde{\bolda}$. Substituting these into (\ref{kgene}) we obtain
\begin{equation*}
S^{-1}D_{ab}D_{bb}^{-1} D_{ba}S^{\prime -1} \hat{\bolda} = \rho^2 \hat{\bolda}.
\end{equation*}
If regularisation using the parameter $\kappa$ is combined with dimensionality reduction the problem becomes 
\begin{equation}\label{hardoon_standard}
S^{-1}D_{ab}\big(D_{bb} + \kappa I \big)^{-1} D_{ba}S^{\prime -1} \hat{\bolda} = \rho^2 \hat{\bolda}.
\end{equation} 
A numerical example of the method presented by \cite{hardoon2004canonical} is given in Example \ref{hardoon_kcca}.

\begin{example}\label{hardoon_kcca} We generate a simulated dataset as follows. The data matrices $X_a$ and $X_b$ of sizes $n \times p$ and $n \times q$, where $n=10000$, $p=7$ and $q=8$, respectively as follows. The seven variables of $X_a$ are generated from a random univariate normal distribution, $\mathbf{a}_1, \mathbf{a}_2, \dots, \mathbf{a}_7 \sim N(0,1)$. We generate the following relations 
\begin{align*}
\mathbf{b}_1 &= \exp(\mathbf{a}_3) + \boldsymbol \xi_1 \\
\mathbf{b}_2 &= \mathbf{a}_1^3 + \boldsymbol \xi_2 \\
\mathbf{b}_3 &= -\mathbf{a}_4 + \boldsymbol \xi_3
\end{align*}
where $\boldsymbol \xi_1 \sim N(0,0.4)$, $\boldsymbol \xi_2 \sim N(0,0.2)$ and $\boldsymbol \xi_3 \sim N(0,0.3)$ denote vectors of normal noise. The five other variables of $X_b$ are generated from a random univariate normal distribution, $\mathbf{b}_4, \mathbf{b}_5, \dots, \mathbf{b}_8 \sim N(0,1)$. The data is standardised such that every variable has zero mean and unit variance.

A Gaussian kernel is used for both views. The width parameter is set using the median trick to $\sigma_a = 3.56$ and $\sigma_b = 3.60.$ The kernels were centred. The positions $\bolda$ and $\boldb$ are found solving the standard eigenvalue problem in (\ref{hardoon_standard}) and applying the Equation (\ref{hardoon_beta}). We set the regularisation parameter $\kappa=0.5$.

The first three canonical correlations at the optimal parameter values were $\langle \za^1,\zb^1 \rangle=0.97$, $\langle \za^2,\zb^2 \rangle=0.97$, and $\langle \za^3,\zb^3 \rangle=0.96.$ The correlation coefficients between the simulated relations and the transformed variables are shown in Table \ref{tab:two}. The exponential relation was extracted in the first pair $(\za^1,\zb^1)$, the 3$^{rd}$ order polynomial relation was extracted in the second pair $(\za^2,\zb^2)$ and the linear relation in the third pair $(\za^3,\zb^3)$.
\begin{table}%
\tbl{Extracted relations by kernel CCA \label{tab:two}}{%
\begin{tabular}{|c|c|c|c|}
\hline
  & $\za^1$ & $\za^2$ & $\za^3$ \\\hline
$exp(\mathbf{a}_3)$  & \bf{0.91} & 0.01 & 0.05  \\\hline
$\mathbf{a}_1^3$  & 0.01 & \bf{0.92} & 0.04  \\\hline
$-\mathbf{a}_4$  & 0.06 & 0.03 & \bf{0.99}  \\\hline
  & $\zb^1$ & $\zb^2$ & $\zb^3$ \\\hline
$\mathbf{b}_1$  & \bf{0.91} & 0.01 & 0.04  \\\hline
$\mathbf{b}_2$   & 0.01 & \bf{0.94} & 0.05  \\\hline
$\mathbf{b}_3$  & 0.07 & 0.04 & \bf{0.99} \\\hline
\end{tabular}}
\end{table}%
$\qed$
\end{example}

Non-linear relations are also taken into account through neural networks which are employed in deep CCA \cite{andrew2013deep}. In deep CCA, every observation $\mathbf{x}_a^k \in \mathbb{R}^p$ and $\mathbf{x}_b^k \in \mathbb{R}^q$ for $k=1,2,\dots,n$ is non-linearly transformed multiple times in an iterative manner through a layered network. The number of units in a layer determines the dimension of the output vector which is put in the next layer. As is explained in \cite{andrew2013deep}, let the first layer have $c_1$ units and the final layer $o$ units. The output vector of the first layer for the observation $\mathbf{x}_a^1 \in \mathbb{R}^p$, is $\mathbf{h}_1=s(S_1^1 \mathbf{x}_a^1 + b_1^1) \in \mathbb{R}^{c_1}$, where $S_1^1 \in \mathbb{R}^{c_1 \times p}$ is a matrix of weights, $b_1^1 \in \mathbb{R}^{c_1}$ is a vector of bias, and $s: \mathbb{R} \mapsto \mathbb{R}$ is a non-linear function applied to each element. The logistic and tanh functions are examples of popular non-linear functions. The output vector $\mathbf{h}_1$ is then used to compute the output of the following layer in similar manner. The final transformed vector $f_1(\mathbf{x}_a^1)=s(S_d^1 h_{d-1}+b_d^1)$ is in the space of $\mathbb{R}^{o}$, for a network with $d$ layers. The same procedure is applied to the observations $\mathbf{x}_b^k \in \mathbb{R}^q$ for $k=1,2,\dots,n$. 

In deep CCA, the aim is to learn the optimal parameters $S_d$ and $b_d$ for both views such that the correlation between the transformed observations is maximised. Let $H_a \in \mathbb{R}^{o \times n}$ and $H_b \in \mathbb{R}^{o \times n}$ denote the matrices that have the final transformed output vectors in their columns. Let $\tilde{H_a}=H_a-\frac{1}{n}H_a\mathbf{1}$ denote the centered data matrix and let $\hat{C}_{ab}=\frac{1}{m-1}\tilde{H}_a\tilde{H}_{b}^T$ and $\hat{C}_{aa}=\frac{1}{m-1}\tilde{H}_a\tilde{H}_{a}^T+r_a I$, where $r_a$ is a regularisation constant, denote the covariance and variance matrices. Same formulae are used to compute the covariance and variance matrices for view $b$. As in section \ref{basic}, the total correlation of the top $k$ components of $H_a$ and $H_b$ is the sum of the top $k$ singular values of the matrix $T=\hat{C}_{aa}^{-1/2}\hat{C}_{ab}\hat{C}_{bb}^{-1/2}$. If $k=o$, the correlation is given by the trace norm of $T$, that is $$corr(H_a,H_b)=tr(T^T T)^{1/2}.$$ The optimal parameters $S_d$ and $b_d$ maximise the trace norm using gradient-based optimisation. The details of the algorithm can be found in \cite{andrew2013deep}.

In summary, kernel and deep CCA provide alternatives to the linear CCA when the relations in the data can be considered to be non-linear and the sample size is small in relation to the data dimensionality. When applying kernel CCA on a real dataset, prior knowledge of the relations of interest can help in the analysis of the results. If the data is assumed to contain both linear and non-linear relations a Gaussian kernel could be a first option. The choice of the kernel function depends on what kind of relations the data can be considered to contain. The possible relations can be extracted by testing how the image pairs correlate with the functions of variables. Deep CCA provides an alternative to compute maximal correlation between the views although the neural network makes the identification of the type of relations difficult.

\subsection{Improving the Interpretability by Enforcing Sparsity}

The extraction of the linear relations between the variables in CCA and regularised CCA relies on the values of the entries of the position vectors that have images on the unit ball with a minimum enclosing angle. The relations can be inferred when the number of variables is not too large for a human to interpret. However, in modern data analysis, it is common that the number of variables is of the order of tens of thousands. In this case, the values of the entries of the position vectors should be constrained such that only a subset of the variables would have a non-zero value. This would facilitate the interpretation since only a fraction of the total number of variables need to be considered when inferring the relations. 

To constrain some of the values of the entries of the position vectors to zero, which is also referred to as to enforce sparsity, tools of convex analysis can be applied. In literature, sparsity has been enforced on the position vectors using soft-thresholding operators \cite{parkhomenko2007genome}, elastic net regularisation \cite{waaijenborg2008quantifying}, penalised matrix decomposition combined with soft-thresholding \cite{witten2009penalized}, and convex least squares optimisation \cite{hardoon2011sparse}. The sparse CCA formulations presented in \cite{parkhomenko2007genome,waaijenborg2008quantifying,witten2009penalized} find sparse position vectors that can be applied to infer linear relations between the variables with non-zero entries. The formulation in \cite{hardoon2011sparse} differs from the preceding propositions in terms of the optimisation criterion. The canonical correlation is found between the image obtained from the linear transformation defined by the data space of one view and the image obtained from the linear transformation defined by the kernel of the other view. The selection of which sparse CCA should be applied for a specific task depends on the research question and prior knowledge regarding the variables. 

The sparse CCA algorithm of \cite{parkhomenko2007genome} can be applied when the aim is to find sparse position vectors and no prior knowledge regarding the variables is available. The positions and images are solved using the SVD, as presented in Section \ref{solving}. Sparsity is enforced on the entries of the positions by iteratively applying the soft-thresholding operator \cite{donoho1995adapting} on the pair of left and right orthonormal singular vectors until convergence. The soft-thresholding operator is a proximal mapping of the $L_1$ norm \cite{bach2011convex}. The consecutive pairs of sparse left and right singular vectors are obtained by deflating the found pattern from the matrix on which the SVD is computed. The sparse CCA hence results in a sparse set of linearly related variables.

The elastic net CCA \cite{waaijenborg2008quantifying} finds sparse position vectors but also considers possible groupings in the variables. The elastic net \cite{zou2005regularization} combines the LASSO \cite{tibshirani1996regression} and the ridge \cite{hoerl1970ridge} penalties. The elastic net penalty incorporates a grouping effect in the variable selection. The term variable selection refers to that a selected variable has a non-zero entry in the position vector. In the soft-thresholding CCA of \cite{parkhomenko2007genome}, the assignment of a non-zero entry is independent of the other entries within the vector. In the elastic net CCA, the ridge penalty groups the variables by the values of the entries and the LASSO penalty either eliminates a group by shrinking the entries of the variables within the group to zero or leaves them as non-zero. The algorithm is based on an iterative scheme of multiple regression. As in \cite{parkhomenko2007genome}, the computations are performed in the data space and therefore the extracted relations are also linear.

The penalised matrix decomposition (PMD) formulation of sparse CCA \cite{witten2009penalized} is based on finding low-rank approximations of the covariance matrix $C_{ab}$. An $n \times p$ sized matrix $X$ with rank $K < min(p,q)$ can be approximated using the SVD \cite{eckart1936approximation} by
\begin{equation*}
\sum_{k=1}^r \sigma_k \mathbf{u}_k \mathbf{v}_k^T = \underset{\tilde{X} \in M(r)}{\text{argmin}} ||X-\tilde{X}||_F^2
\end{equation*}
where $\mathbf{u}_k$ denotes the column of the matrix $U$, $\mathbf{v}_k$ denotes the column of the matrix $U$, $\sigma_k$ denotes the $k^{th}$ singular value on the diagonal of $S$, $M(r)$ is the set of rank $r$ $n \times p$ matrices and $r<<K$. In the case of CCA, the matrix to be approximated is the covariance matrix  $X=C_{ab}.$ The optimisation problem in the PMD context is given by
\begin{gather*}
 \min_{\wa \in \mathbb{R}^p, \wb \in \mathbb{R}^q} \frac{1}{2} || C_{ab} - \sigma \wa \wb^T||_F^2, \\
||\wa||_2 = 1 \quad ||\wb||_2 = 1, \\
||\wa||_1 \leq c_1 \quad ||\wb||_1 \leq c_2, \quad \sigma \geq 0
\end{gather*}
which is equivalent to
\begin{gather*}
\cos \theta = \max_{\wa \in \mathbb{R}^p, \wb \in \mathbb{R}^q} \wa^T C_{ab} \wb, \\
||\wa||_2 \leq 1 \quad ||\wb||_2 \leq 1, \\
||\wa||_1 \leq c_1 \quad ||\wb||_1 \leq c_2.
\end{gather*}
The aim is to find $r$ pairs of sparse position vectors $\wa$ and $\wb$ such that their outer products represent low-rank approximations of the original $C_{ab}$ and hence extracts the $r$ linear relations from the data.

The exact derivation of the algorithm to solve the PMD optimisation problem is given in \cite{witten2009penalized}. In general, the position vectors, that generate 1-rank approximations of the covariance matrix, are found in an iterative manner. To find one 1-rank approximation, the soft-thresholding operator is applied as follows. Let the soft-thresholding operator be given by
\begin{equation*}
S(a,c) = sign(a)(|a|-c)_{+}
\end{equation*}
where $c>0$ is a constant. The following formula is applied in the derivation of the algorithm
\begin{gather*}
\max_{\mathbf{u}} \langle \mathbf{u}, \mathbf{a} \rangle, \\
s.t. \quad ||\mathbf{u}||_2^2 \leq 1, ||\mathbf{u}||_1 < c.
\end{gather*}
The solution is given by $\mathbf{u}= \frac{S(\mathbf{a},\delta)}{||S(\mathbf{a},\delta)||_2}$ with $\delta=0$ if $||\mathbf{u}_1|| \leq c$. Otherwise, $\delta$ is selected such that  $||\mathbf{u}_1|| = c.$ Sparse position vectors are the obtained by Algorithm \ref{alg:pmd}. At every iteration, the $\delta_1$ and $\delta_2$ are selected by binary search.
\begin{algorithm}[]
\SetAlgoNoLine
Initialise $||\mathbf{\wb}||_2=1$ \;
\Repeat{convergence }{
        $\mathbf{\wa} \leftarrow \frac{S(C_{ab}\mathbf{\wb},\delta_1)}{||S(C_{ab}\mathbf{\wb},\delta_1)||_2}$ where $\delta_1=0$ if $||\wa||_1 \leq c_1$, otherwise $\delta_1$ is chosen such that $||\wa||_1 = c_1$ and $c_1 >0$ \;
                $\mathbf{\wb} \leftarrow \frac{S(C_{ab}^T\mathbf{\wa},\delta_2)}{||S(C_{ab}^T\mathbf{\wa},\delta_2)||_2}$ where $\delta_2=0$ if $||\wb||_1 \leq c_2$, otherwise $\delta_2$ is chosen such that $||\wb||_1 = c_2$ and $c_2 >0$ \;
      }
      $\sigma \leftarrow \wa^T C_{ab} \wb$
\caption{Computation of a 1-rank approximation of the covariance matrix}
\label{alg:pmd}
\end{algorithm}
To obtain several 1-rank approximations, a deflation step is included such that when the converged vectors $\wa$ and $\wb$ are found, the extracted relations are subtracted from the covariance matrix $C_{ab}^{k+1} \leftarrow C^{k}-\sigma_k \wa^k \wb^{k T}.$ In this way, the successive solutions remain orthogonal which is a contstraint of CCA.

\begin{example}\label{pmd} To demonstrate the PMD formulation of sparse CCA, we generate the following data. The data matrices $X_a$ and $X_b$ of sizes $n \times p$ and $n \times q$, where $n=50$, $p=100$ and $q=150$, respectively as follows. The variables of $X_a$ are generated from a random univariate normal distribution, $\mathbf{a}_1, \mathbf{a}_2, \cdots, \mathbf{a}_{100} \sim N(0,1)$. We generate the following linear relations 
\begin{align}
\mathbf{b}_1 &= \mathbf{a}_3 + \boldsymbol \xi_1 \label{rel1}\\
\mathbf{b}_2 &= \mathbf{a}_1 + \boldsymbol \xi_2 \label{rel2}\\
\mathbf{b}_3 &= -\mathbf{a}_4 + \boldsymbol \xi_3 \label{rel3}
\end{align}
where $\boldsymbol \xi_1 \sim N(0,0.08), \boldsymbol \xi_2 \sim N(0,0.07),$ and $ \boldsymbol \xi_3 \sim N(0,0.05)$ denote vectors of normal noise. The other variables of $X_b$ are generated from a random univariate normal distribution, $\mathbf{b}_4, \mathbf{b}_5, \cdots, \mathbf{b}_{150} \sim N(0,1)$. The data is standardised such that every variable has zero mean and unit variance.

We apply the R implementation of \cite{witten2009penalized} which is available in the PMA package. We extract three rank-1 approximations. The values of the entries of the pairs of position vectors $(\wa^1,\wb^1),(\wa^2,\wb^2)$ and $(\wa^3,\wb^3)$ corresponding to canonical correlations $\langle \za^1,\zb^1 \rangle = 0.95$, $\langle \za^2,\zb^2 \rangle = 0.92$, $\langle \za^3,\zb^3 \rangle = 0.91$ are shown in Figure \ref{pmdfig}. The first 1-rank approximation extracted (\ref{rel3}), the second (\ref{rel2}), and the third (\ref{rel3}). 
\begin{figure}
\centerline{\includegraphics[trim={1cm 1cm 1.5cm 1cm},clip,width=0.78\textwidth]{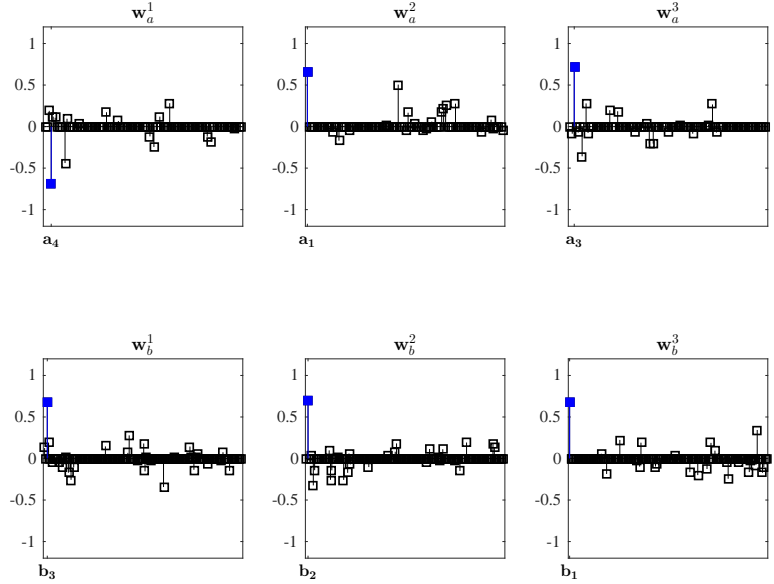}}
\caption{The values of the entries of the position vector pairs $(\wa^1,\wb^1),(\wa^2,\wb^2)$ and $(\wa^3,\wb^3)$ obtained using the PMD method for sparse CCA are shown. The entry of maximum absolute value is coloured blue. The negative linear relation between $\mathbf{a}_4$ and $\mathbf{b}_3$ is extracted in the first 1-rank approximation. The positive linear relations between $\mathbf{a}_1$ and $\mathbf{b}_2$ and $\mathbf{a}_3$ and $\mathbf{b}_1$ are extracted in the second and third 1-rank approximations.}
\label{pmdfig}
\end{figure}
$\qed$
\end{example}

The sparse CCA of \cite{hardoon2011sparse} is a sparse convex least squares formulation that differs from the preceding versions. The canonical correlation is found between the linear transformations between a data space view and a kernel space view. The aim is to find a sparse set of variables in the data space view that relate to a sparse set of observations, represented in terms of relative similarities, in the kernel space view. An example of a setting, where relations of this type can provide useful information, is in bilingual analysis as described in \cite{hardoon2011sparse}. When finding relations between words of two languages, it may be useful to know in what kind of contexts can a word be used in the translated language. The optimisation problem is given by
\begin{gather*}
\cos(\za,\zb) = \max_{\za, \zb \in \mathbb{R}^n} \langle \za,\zb \rangle = \wa^T X_a^T K_b \boldb, \\
 ||\za||_2=\sqrt{\wa^T X_a^T X_a \wa} = 1 \quad ||\zb||_2=\sqrt{\boldb^T K_b^2 \boldb} = 1
\end{gather*}
which is equivalent to the convex sparse least squares problem
\begin{gather}
\min_{\wa,\boldb} ||X_a \wa - K_b \boldb||^2 + \mu ||\wa||_1 + \gamma ||\tilde{\boldb}||_1 \\
s.t \quad ||\boldb||_{\infty}=1
\end{gather}
where $\mu$ and $\gamma$ are fixed parameters that control the trade-off between function objective and the level of sparsity of the entries of the position vectors $\wa$ and $\boldb$. The constraint $||\boldb||_{\infty}=1$ is set to avoid the trivial solution ($\wa=\mathbf{0},\boldb=\mathbf{0}$). The $k^{th}$ entry of $\boldb$ is set to $\beta_k=1$ and the remaining entries in $\tilde{\boldb}$ are constrained by 1-norm. The idea is to fix one sample as a basis for comparison and rank the other similar samples based on the fixed sample. The optimisation problem is solved by iteratively minimising the gap between the primal and dual Lagrangian solutions. The procedure is outlined in Algorithm \ref{alg:scca}. The exact computational steps can be found in \cite{hardoon2011sparse}.
\begin{algorithm}[t]
\SetAlgoNoLine
\Repeat{convergence }{
        1. Use the dual Lagrangian variables to solve the primal variables \;
        2. Check whether all constraints on the primal variables hold \;
        3. Use the primal variables to solve the dual Lagrangian variables \;
        4. Check whether all dual Lagrangian variable constraints hold \;
        5. Check whether 2 holds, if not go to 1 \;
      }
\caption{Pseudo-code to solve the convex sparse least squares problem }
\label{alg:scca}
\end{algorithm}
The Algorithm \ref{alg:scca} is used to extract one relation or pattern from the data. To extract the successive patterns, deflation is applied to obtain the residual matrices from which the already found pattern is removed. In Example \ref{scca_ex}, the extraction of the first pattern is shown.

\begin{example}\label{scca_ex} In the sparse CCA of \cite{hardoon2011sparse}, the idea is to determine the relations of the variables in the data space view $X_a$ to the observations in the kernel space view $K_b$ where the observations comprise the variables of the view $b$. This setting differs from all of the previous examples where the idea was to find relations between the variables. Since one of the views is kernelised, the relations cannot be explicitly simulated. We therefore demonstrate the procedure on data generated from random univariate normal distribution as follows. The data matrices $X_a$ and $X_b$ of sizes $n \times p$ and $n \times q$, where $n=50$, $p=100$ and $q=150$, respectively are generated as follows. The variables of $X_a$ and $X_b$ are generated from random univariate normal distribution, $\mathbf{a}_1, \mathbf{a}_2, \cdots, \mathbf{a}_{100} \sim N(0,1)$ and $\mathbf{b}_1, \mathbf{b}_2, \cdots, \mathbf{b}_{150} \sim N(0,1)$ respectively. The data is standardised such that every variable has zero mean and unit variance. 

The Gaussian kernel function $K(\mathbf{x},\mathbf{y})=exp(-||\mathbf{x}-\mathbf{y}||^2/2\sigma^2)$  is used to compute the similarities for the view $b$. The choice of the kernel is justified since the underlying distribution is normal. The width parameter is set to $\sigma=17.25$ using the median trick. The kernel matrix is centred by $\tilde{K} = K - \frac{1}{n} \mathbf{j} \mathbf{j}^T K - \frac{1}{n} K \mathbf{j} \mathbf{j}^T  + \frac{1}{n^2} (\mathbf{j}^T K\mathbf{j}) \mathbf{j} \mathbf{j}^T$ where $\mathbf{j}$ contains only entries of value one \cite{shawe2004kernel}. 

To find the positions $\wa$ and $\boldb$, we solve
\begin{gather*}
f = \min_{\wa,\boldb} ||X_a \wa - K_b \boldb||^2 + \mu ||\wa||_1 + \gamma ||\tilde{\boldb}||_1 \\
s.t \quad ||\boldb||_{\infty}=1
\end{gather*}
using the implementation proposed in \cite{uurtio2015canonical}. As stated in \cite{hardoon2011sparse}, to determine which variable in the data space view $X_a$ is most related to the observation in $K_b$, the algorithm needs to be run for all possible values of $k$. This means that every observation is in turn set as a basis for comparison and a sparse set of the remaining observations $\tilde{\boldb}$ is computed. The optimal value of $k$ gives the minimum objective value $f$. 

We run the algorithm by initially setting the value of the entry $\beta_k=1$ for $k=1,2,\dots,n$. The minimum objective value $f=0.03$ was obtained at $k=29$. This corresponds to a canonical correlation of $\langle \za,\zb \rangle = 0.88$. The values of the entries of $\wa$ and $\boldb$ are shown in Figure \ref{scca_exfig}. The observation corresponding to $k=29$ in the kernelised view $K_b$ is most related to the variables $\mathbf{a}_{15}, \mathbf{a}_{16}, \mathbf{a}_{18}, \mathbf{a}_{20},$ and $\mathbf{a}_{24}$. 
\begin{figure}
\centerline{\includegraphics[trim={1cm 3cm 1.5cm 3cm},clip,width=0.75\textwidth]{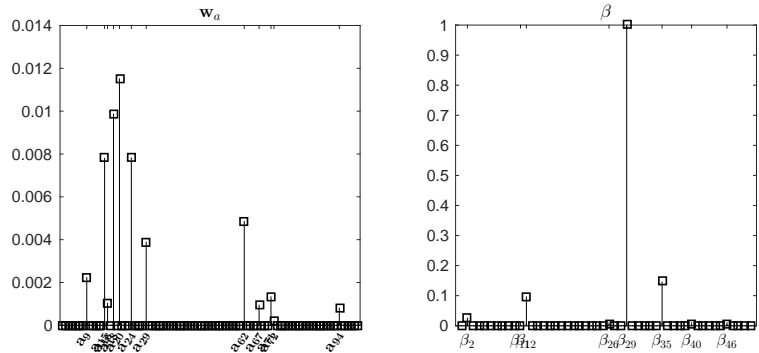}}
\caption{The values of the entries of the positions $\wa$ and $\boldb$ at the optimal value of $k$ are shown.}
\label{scca_exfig}
\end{figure}
$\qed$
\end{example}

The sparse versions of CCA can be applied to settings when the large number of variables hinders the inference of the relations. When the interest is to extract sparse linear relations between the variables, the proposed algorithms of \cite{parkhomenko2007genome,waaijenborg2008quantifying,witten2009penalized} provide a solution. The algorithm of \cite{hardoon2011sparse} can be applied if the focus is to find how the variables of one view relate to the observations that correspond to the combined sets of the variables in the other view. In other words, the approach is useful if the focus is not to uncover the explicit relations between the variables but to gain insight how a variable relates to a complete set of variables of an observation.

\section{Discussion}

This tutorial presented an overview on the methodological evolution of canonical correlation methods focusing on the original linear, regularised, kernel, and sparse CCA. Succinct reviews were also conducted on the Bayesian and neural network-based deep CCA variants. The aim was to explain the theoretical foundations of the variants using the linear algebraic interpretation of CCA. The methods to solve the CCA problems were described using numerical examples. Additionally, techniques to assess the statistical significance of the extracted relations and the generalisability of the patterns were explained. The aim was to delineate the applicabilities of the different CCA variants in relation to the properties of the data. 

In CCA, the aim is to determine linear relations between variables belonging to two sets. From a linear algebraic point of view, the relations can be found by analysing the linear transformations defined by the two views of the data. The most distinct relations are obtained by analysing the entries of the first pair of position vectors in the two data spaces that are mapped onto a unit ball such that their images have a minimum enclosing angle. The less distinct relations can be identified from the successive pairs of position vectors that correspond to the images with a minimum enclosing angle obtained from the orthogonal complements of the preceding pairs of images. This tutorial presented three standard ways of solving the CCA problem, that is by solving either a standard \cite{hotelling1935most,hotelling1936relations} or a generalised eigenvalue problem \cite{bach2002kernel,hardoon2004canonical}, or by applying the SVD \cite{healy1957rotation,ewerbring1989canonical}.

The position vectors of the two data spaces, that convey the related pairs of variables, can be obtained using alternative techniques than the ones selected for this tutorial. The three methods were chosen because they have been much applied in CCA literature and they are relatively straightforward to explain and implement. Additionally, to understand the further extensions of CCA, it is important to know how it originally has been solved. The extensions are often further developed versions of the standard techniques. 

For didactic purposes, the synthetic datasets used for the worked examples were designed to represent optimal data settings for the particular CCA variants to uncover the relations. The relations were generated to be one-to-one, in other words one variable in one view was related with only one variable in the other view. In real datasets, which are often much larger than the synthetic ones in this paper, the relations may not be one-to-one but rather many-to-many (one-to-two, two-to-three, etc.). As in the worked examples, these relations can also be inferred by examining the entries of the position vectors of the two data spaces. However, the understanding of how the one-to-one relations are extracted provides means to uncover the more complex relations.  

To apply the linear CCA, the sample size needs to exceed the number of variables of both views which means that the system is required to be overdetermined. This is to guarantee the non-singularity of the variance matrices. If the sample size is not sufficient, regularisation \cite{vinod1976canonical} or Bayesian CCA \cite{klami2013bayesian} can be applied. The feasibility of regularisation has not been studied in relation to the number of variables exceeding the number of observations. Improving the invertibility by introducing additional bias has been shown to work in various settings but the limit when the system is too underdetermined that regularisation cannot assist in recovering the underlying relations has not been resolved. Bayesian CCA is more robust against outlying observations, when compared with linear CCA, due to its generative model structure.

In addition to linear relations, non-linear relations are taken into account in kernelised and neural network-based CCA. Kernel methods enable the extraction of non-linear relations through the mapping to a Hilbert space \cite{bach2002kernel,hardoon2004canonical}. When applying kernel methods in CCA, the disparity between the number of observations and variables can be huge due to very dimensional kernel induced feature spaces, a challenge that is tackled by regularisation. The types of relations that can be extracted, is  determined by the kernel function that was selected for the mapping. Linear relations are extracted by a linear kernel and non-linear relations by non-linear kernel functions such as the Gaussian kernel. Although kernelisation extends the range of extractable relations, it also complicates the identification of the type of relation. A method to determine the type of relation involves testing how the image vectors correlate with a certain type of function. However, this may be difficult if no prior knowledge of the relations is available. Further research on how to select the optimal kernel functions to determine the most distinct relations underlying in the data could facilitate the final inference making. 
Neural network-based deep CCA is an alternative to kernelised CCA, when the aim is to find a high correlation between the final output vectors obtained through multiple non-linear transformations. However, due to the network structure, it is not straightforward to identify the relations between the variables.  

As a final branch of the CCA evolution, this tutorial covered sparse versions of CCA. Sparse CCA variants have been developed to facilitate the extraction of the relations when the data dimensionality is too high for human interpretation. This has been addressed by enforcing sparsity on the entries of the position vectors \cite{parkhomenko2007genome,waaijenborg2008quantifying,witten2009penalized}. As an alternative to operating in the data spaces, \cite{hardoon2011sparse} proposed a primal-dual sparse CCA in which the relations are obtained between the variables of one view and observations of the other. The sparse variants of CCA in this tutorial were selected based on how much they have been applied in literature. As a limitation of the selected variants, sparsity is enforced on the entries of the position vectors without regarding the possible underlying dependencies between the variables which has been addressed in the literature of structured sparsity \cite{chen2012structured}.

In addition to studying the techniques of solving the optimisation problems of CCA variants, this tutorial gave a brief introduction to evaluating the canonical correlation model. Bartlett's sequential test procedure \cite{bartlett1938further,bartlett1941statistical} was given as an example of a standard method to assess the statistical significance of the canonical correlations. The techniques of identifying the related variables through visual inspection of biplots \cite{meredith1964canonical,ter1990interpreting} were presented. To assess whether the extracted relations can be considered to occur in any data with the same underlying sampling distribution, the method of applying both training and test data was explained. As an alternative method, the statistical significance of the canonical correlation model could be assessed using permutation tests \cite{rousu2013biomarker}. The visualisation of the results using the biplots is mainly applicable in the case of linear relations. Alternative approaches could be considered to visualise the non-linear relations extracted by kernel CCA. 

To conclude, this tutorial compiled the original, regularised, kernel, and sparse CCA into a unified framework to emphasise the applicabilities of the four variants in different data settings. The work highlights which CCA variant is most applicable depending on the sample size, data dimensionality and the type of relations of interest. Techniques for extracting the relations are also presented. Additionally, the importance of assessing the statistical significance and generalisability of the relations is emphasised. The tutorial hopefully advances both the practice of CCA variants in data analysis and further development of novel extensions. 

The software used to produce the examples in this paper are available for download at
https://github.com/aalto-ics-kepaco/cca-tutorial.

\begin{acks}
The work by Viivi Uurtio and Juho Rousu has been supported in part by Academy of Finland (grant 295496/D4Health). Jo\~ao M. Monteiro was supported by a PhD studentship awarded by Funda\c{c}\~ao para a Ci\^encia e a Tecnologia (SFRH/BD/88345/2012). John Shawe-Taylor acknowledges the support of the EPSRC through the C-PLACID project Reference: EP/M006093/1.
\end{acks}

\bibliographystyle{ACM-Reference-Format-Journals}
\bibliography{csur_refs}

\received{February 2017}{July 2017}{August 2017}





\end{document}